\PassOptionsToPackage{svgnames}{xcolor}
\documentclass{article} %
\usepackage{iclr2027_conference,times}

\usepackage{amsmath,amsfonts,bm}

\def\eqref#1{equation~\ref{#1}}

\def\1{\bm{1}}

\DeclareMathAlphabet{\mathsfit}{\encodingdefault}{\sfdefault}{m}{sl}
\SetMathAlphabet{\mathsfit}{bold}{\encodingdefault}{\sfdefault}{bx}{n}

\usepackage{hyperref}
\usepackage{url}

\usepackage[utf8]{inputenc} %
\usepackage[T1]{fontenc}    %
\usepackage{booktabs}       %
\usepackage{amsfonts}       %
\usepackage{nicefrac}       %
\usepackage{microtype}      %
\usepackage{xcolor}         %

\usepackage{amsmath}
\usepackage{amssymb}
\usepackage{mathtools}
\usepackage{amsthm}

\usepackage{enumitem}
\usepackage{titlesec}

\usepackage{graphicx}
\usepackage{subcaption}
\usepackage{wrapfig}
\usepackage{multirow}
\usepackage{array}
\usepackage{tabularx}

\usepackage[capitalize,noabbrev]{cleveref}

\definecolor{linkblue}{RGB}{0,60,120}

\hypersetup{
    colorlinks=true,
    linkcolor=linkblue,  %
    citecolor=linkblue,  %
    urlcolor=linkblue    %
}

\titlespacing*{\section}
  {0pt}{1.2ex plus .2ex minus .2ex}{.7ex plus .1ex}
\titlespacing*{\subsection}
  {0pt}{1.0ex plus .2ex minus .2ex}{.4ex plus .1ex}
\titlespacing*{\subsubsection}
  {0pt}{.8ex plus .2ex minus .1ex}{.3ex plus .1ex}

\title{Growing to See: Efficient Pretraining of Self-Supervised Vision Transformers}

\author{
    \parbox{\linewidth}{\centering
        Amir Mohammad Karimi Mamaghan$^1$\thanks{
        Equal contribution.\\
        Correspondence to Amir Mohammad Karimi Mamaghan
        (\texttt{amkm@kth.se}) and Ferdinand Kapl
        (\texttt{ferdinand.kapl@tum.de}).}\quad
        Ferdinand Kapl$^{2,3,4}$\footnotemark[1] \\[0.4em]
        Andrea Dittadi$^{2,3}$\quad
        Karl Henrik Johansson$^{1,6}$\quad
        Kaitlin Maile$^5$ \\[0.4em]
        Johannes von Oswald$^5$\quad
        Stefan Bauer$^{2,3,4}$ \\[0.8em]
        \normalfont
        $^1$KTH Royal Institute of Technology \\
        $^2$Helmholtz AI, Munich\quad
        $^3$Technical University of Munich \\
        $^4$Munich Center for Machine Learning (MCML) \\
        $^5$Google, Paradigms of Intelligence Team \\
        $^6$Digital Futures
    }
}

\iclrarxivcopy
\begin{document}

\maketitle

\begin{abstract}

Progressive depth growth builds large networks by starting with shallow models and inserting layers during training. Recent advances in large language models have renewed interest in this longstanding approach, demonstrating benefits for training efficiency and reasoning alongside changes in internal computations. However, how growing affects representation quality and computational patterns in self-supervised vision transformers remains unclear.
In this work, we systematically investigate progressive depth growth in two popular self-supervised families, namely masked autoencoders (MAE) and DINOv3, and analyze how growing affects downstream performance, intermediate representations, and internal computations. 
Across global and dense tasks, we show that growing improves downstream performance while reducing training FLOPs by 13.9--24.8\%, with ImageNet linear-probe accuracy gains of up to 3.36 percentage points. Growing also improves downstream performance at intermediate layers and reshapes internal computational patterns, introducing recurring alternations between global and localized attention in MAE and earlier local-to-global attention progression in self-distilled DINOv3.
Together, these findings establish progressive depth growth as an effective and efficient pretraining strategy for improving downstream performance across global and dense visual tasks with less computation, and advance our understanding of how growing shapes visual computation.\looseness=-1

\end{abstract}

\section{Introduction}
\label{sec:introduction}

Recent work on large language models (LLMs) has renewed interest in a longstanding idea in deep learning: start with a smaller network and gradually expand it during training \citep{fahlman1989cascade,chen2015net2net,reddi2023efficient}. Progressive depth growth implements this idea by starting with a shallow network and adding layers, reducing pretraining compute by allocating part of the training budget to smaller models. Beyond efficiency, growing and related approaches such as looping have been connected to improved reasoning and changes in learned computation \citep{saunshi2024inductive,saunshi2025latent,kapl2026do,kapl2026from}.
These findings suggest that growing can influence more than the cost of training: introducing layers progressively may shape the representations and emerging computational patterns.\looseness=-1

Vision transformers (ViTs) \citep{dosovitskiy2020image} provide a compelling setting for investigating these effects. Their representations support tasks ranging from global image-level classification to dense pixel-level prediction, allowing us to examine how growing affects the emergence of information at different granularities. Analyses of pretrained ViTs have identified transitions between local and global processing and repeated or phase-like behavior across layers \citep{park2023what,jacobs2026block}. Other studies show that attention and representations evolve differently across self-supervised learning (SSL) objectives \citep{park2023what,shekhar2023objectives}.
These observations raise a natural question: how does this organization change when the architecture grows during training?\looseness=-1

Despite these developments, how progressive growth shapes representations and computational patterns in self-supervised vision models remains insufficiently understood. Existing growth methods in vision primarily target training efficiency, focusing largely on supervised ViTs \citep{li2022automated,ding2023network,wang2023learning,fu2023triple} or image--text models \citep{deng2023growclip,fan2026chain}, with limited analysis of how growing changes internal computations. 
This question is particularly interesting across popular SSL objectives: Masked Image Modeling (MIM) \citep{bao2021beit,he2022masked} and self-distillation \citep{caron2021emerging,oquab2023dinov2}, whose different training signals may lead to different responses to progressive growth.

In this work, we study progressive depth growth in two representative self-supervised model families: masked autoencoders (MAE) \citep{he2022masked} and DINOv3 \citep{simeoni2025dinov3}. Starting from shallow networks, we progressively insert layers by copying existing blocks. We evaluate Stacking, MIDAS, and LIDAS across model sizes, growth schedules, training budgets, and distillation settings, comparing grown models with fixed-depth baselines at matched training steps and data exposure, with identical final architectures. We combine a wide range of tasks, from global image classification to dense semantic segmentation and monocular depth estimation, with layer-wise readouts, attention analyses, and layer-skipping interventions to analyze both representation quality and internal computation. Our main findings are:

\begin{itemize}[topsep=0pt, itemsep=0pt, parsep=0pt, partopsep=0pt]
    \item \textbf{Growing improves downstream performance with less pretraining compute (\cref{subsec:exp_downstream}).}
    Across global and dense tasks, growing improves final-layer downstream performance while reducing pretraining FLOPs by 13.9--24.8\%, with unchanged final architectures and inference costs. ImageNet linear-probe accuracy improves by up to 3.36 percentage points and ADE20K mIoU by up to 1.44 points, while NYUv2 depth-estimation RMSE decreases by up to 0.0538 meters. Based on these results, we identify the following practical growing recipes: Stacking performs best for smaller MAEs, while LIDAS is the strongest strategy for larger MAEs and DINOv3.

    \item \textbf{Growing improves representations in intermediate layers (\cref{subsec:exp_layerwise_downstream}).}
    Frozen intermediate representations yield better downstream performance, often with larger gains than at the final layer. These improvements are particularly pronounced in middle layers, showing that growing makes task-relevant information accessible at earlier depths.

    \item \textbf{Growing changes computational patterns (\cref{subsec:exp_computational}).}
    MAE develops recurring alternations between global and localized attention, aligned with the four-layer growth increments, while self-distilled DINOv3 exhibits earlier local-to-global attention progression. MAE's recurring patterns persist after prolonged full-depth training, even as copied weights become less similar. In both families, skipping the earliest layers produces larger relative changes in representations in subsequent layers than in fixed-depth baselines, whereas skipping many later layers produces smaller changes.
    
\end{itemize}

Overall, this work advances our understanding of how progressive depth growth shapes visual representations and internal computation, providing practical guidance for pretraining more capable self-supervised vision models with less compute.

\section{Related Work}
\label{sec:related_work}

\textbf{Self-supervised Visual Representation Learning.}~
Learning visual representations without manual annotations has been a longstanding goal in computer vision. Early context and rotation prediction tasks \citep{doersch2015unsupervised,gidaris2018unsupervised} preceded contrastive methods such as MoCo and SimCLR \citep{he2020momentum,chen2020simple} and teacher--student methods such as BYOL and DINO \citep{grill2020bootstrap,caron2021emerging}. BEiT and MAE use visual-token prediction and pixel reconstruction, respectively \citep{bao2021beit,he2022masked}. iBOT combines masked prediction with self-distillation \citep{zhou2021ibot}, while DINOv2 scales image- and patch-level objectives with curated data \citep{oquab2023dinov2}. DINOv3 introduces Gram anchoring to preserve dense feature quality during prolonged training \citep{simeoni2025dinov3}.

\textbf{Growing Neural Networks.}~
Growing neural networks during training has a long history as a way to build larger models by reusing what smaller ones have already learned. Net2Net \citep{chen2015net2net} introduced function-preserving expansion, later extended to transformers \citep{gesmundo2023composable}. Recent LLM work demonstrated efficient pretraining through depth-wise stacking \citep{du2024stacking}. Beyond efficiency, MIDAS improved reasoning despite similar or slightly worse perplexity, suggesting a growth-induced inductive bias \citep{saunshi2024inductive}. \citet{kapl2026do} linked middle stacking to increased late-layer utilization and permutable computational blocks, and introduced LIDAS, a layer-wise middle-duplication strategy with further reasoning gains. These findings motivate examining how growing shapes visual representations and computation.

\textbf{Growing in Vision.}~
In vision, prior work has primarily pursued training acceleration through model expansion. AutoProg uses a supernet to search depth and resolution schedules for supervised ViTs \citep{li2022automated}. Network Expansion grows supervised CNNs and ViTs with architecture-specific rules \citep{ding2023network}, while LiGO learns expansion mappings and evaluates supervised ViT classification transfer \citep{wang2023learning}. TripLe combines width expansion and staged depth growth in supervised ViTs, copying weights and optimizer states \citep{fu2023triple}. Related approaches study progressive unfreezing for diffusion fine-tuning and efficient growth of image--text models \citep{li2026efficient,deng2023growclip,fan2026chain}. SCG adaptively grows small self-supervised JEPA encoders in width and depth on synthetic environments \citep{berenz2026successive}. Complementary SSL methods use input curricula or layer freezing \citep{wang2024efficienttrain++,erdogan2025layerlock}. However, these growth studies primarily target training efficiency, largely focus on supervised ViTs or image--text models, and often evaluate smaller-scale models and employ more elaborate growth procedures involving supernet training, architecture search, or learned expansion operators, while offering limited analysis of how growth changes internal computation across SSL objectives.

\section{Progressive Depth Growing of Vision Transformers}
\label{sec:method}

Progressive depth growth starts from a shallow network and inserts additional layers during training until reaching a target final depth. By allocating part of training to smaller models, this approach reduces pretraining FLOPs relative to training at the final depth throughout, for the same number of training steps and batch size. We investigate growth strategies previously studied in language modeling \citep{reddi2023efficient,saunshi2024inductive,kapl2026do} in two self-supervised vision settings: masked image reconstruction with MAE and self-distillation with DINOv3. These objectives provide different training signals, leaving open which growth strategies and schedules yield effective visual representations. We separate two design choices: \emph{which} layers to copy (growing strategy), and \emph{when} to increase depth (growing schedule).

\textbf{Growing Strategies.}~
Let $(L_0,\ldots,L_{N-1})$ denote the ordered Transformer layers before a growth event, using zero-based indices. Each event increases depth from $N$ layers to $N+B$, where $N$ is a multiple of $B$, by copying a contiguous slice of $B$ layers and inserting the copies immediately after their source:\looseness=-1
\begingroup
\setlength{\abovedisplayskip}{4pt}
\setlength{\belowdisplayskip}{4pt}
\setlength{\abovedisplayshortskip}{0pt}
\setlength{\belowdisplayshortskip}{4pt}
\begin{equation}
    \bigl(
    L_0,\ldots,L_{s+B-1},
    \widetilde{L}_{s},\ldots,\widetilde{L}_{s+B-1},
    L_{s+B},\ldots,L_{N-1}
    \bigr),
    \label{eq:growing_layers}
\end{equation}
\endgroup
where $s$ is the first source-layer index and $\widetilde{L}_j$ is initialized with the parameters of $L_j$. The trailing sequence is empty when $s+B=N$. Source and copied layers have separate trainable parameters and are updated jointly during subsequent training, without weight sharing. We compare three strategies that differ in their choice of $s$:
\begingroup
\setlength{\abovedisplayskip}{4pt}
\setlength{\belowdisplayskip}{4pt}
\setlength{\abovedisplayshortskip}{0pt}
\setlength{\belowdisplayshortskip}{4pt}
\begin{equation}
    s=
    \left\{
    \begin{array}{ll}
    N-B, & \text{Stacking},\\
    B\left\lfloor (N/B-1)/2\right\rfloor, & \text{MIDAS},\\
    \left\lceil (N-B)/2\right\rceil, & \text{LIDAS}.
    \end{array}
    \right.
    \label{eq:growing_strategies}
\end{equation}
\endgroup
\textbf{Stacking} copies the final $B$ layers and appends them to the network, following the gradual stacking strategy described by \citet{reddi2023efficient}. \textbf{MIDAS} \citep{saunshi2024inductive} partitions the network into consecutive groups of $B$ layers and duplicates the middle group, choosing the earlier group when there are two central groups. \textbf{LIDAS} \citep{kapl2026do} instead duplicates the central $B$-layer slice without requiring alignment with these group boundaries. See \cref{fig:growing_strategies} for an illustration of growing strategies.\looseness=-1

Unless stated otherwise, models start with four layers and grow in increments of $B=4$ until reaching 24 layers for ViT-L or 32 layers for ViT-H. Width, attention-head count, and patch size remain unchanged. For MAE, we grow only the encoder and keep the decoder fixed. For self-distilled DINOv3, the teacher backbone is reset to the expanded student's backbone weights at each growth event, after which exponential moving average updates resume. In the separate pretrained-teacher distillation setting, only the student grows and the target teacher remains fixed. Exact layer-copy mappings, optimizer-state handling, and teacher updates are detailed in \cref{app:growing_setup}.

\textbf{Growing Schedule.}~
Following \citet{saunshi2024inductive}, we allocate training steps across successive depths using Prop-$\alpha$ schedules. Starting from an initial depth of $B$, a $K$-stage run reaches a final depth of $KB$. For a growth-schedule horizon of $T_g$ steps, stage $i\in{1,\ldots,K}$ has depth $iB$ and a duration proportional to $i^\alpha$. Specifically, we set 
\begingroup
\setlength{\abovedisplayskip}{4pt}
\setlength{\belowdisplayskip}{4pt}
\setlength{\abovedisplayshortskip}{0pt}
\setlength{\belowdisplayshortskip}{4pt}
\begin{equation}
\Delta_i=
\left\lfloor
T_g\frac{i^\alpha}{\sum_{j=1}^{K}j^\alpha}
\right\rfloor
\quad (i<K),
\qquad
\Delta_K=T_g-\sum_{i=1}^{K-1}\Delta_i,
\label{eq}
\end{equation}
\endgroup
assigning the rounding remainder to the final stage. Growth occurs after stages $1,\ldots,K-1$, at steps $t_i=\sum_{j=1}^{i}\Delta_j$. Larger values of $\alpha$ allocate more steps to deeper stages, increasing pretraining FLOPs for a fixed number of training steps. The schedule therefore controls the trade-off between computation saved at shallow depths and training allocated to deeper models. In our experiments, we compare Prop-1, Prop-2, and Prop-3, with strategy and schedule ablations in \cref{app:ablation_growing_schedule_strategy}. The horizon $T_g$ includes training at the final depth and does not need to equal the total training duration. 
Growth transitions do not restart the global learning-rate schedule. Exact stage boundaries and implementation details are provided in \cref{app:growing_setup}.

\section{Experiments}
\label{sec:experiments}

\textbf{Setup.}~
We study progressive depth growth for two popular SSL models: MAE \citep{he2022masked} with ViT-L/16 and ViT-H/14 backbones, and DINOv3 \citep{simeoni2025dinov3} with a ViT-L/16 backbone. All models are trained from scratch on ImageNet-1k \citep{deng2009imagenet} using recipes adapted from their original setup. Unless stated otherwise, MAE is trained for 50k steps ($\sim$160 epochs) and DINOv3 for 125k steps ($\sim$200 epochs). Within each comparison, we match training steps, batch size, and other training hyperparameters, ensuring equal data exposure. Grown models reach the same final architecture and inference cost as their fixed-depth baselines. The main MAE ViT-L and DINOv3 comparisons use cosine learning-rate schedules, while MAE ViT-H uses warmup--stable--decay (WSD). Full training and evaluation details are provided in \cref{app:training_setup,app:evaluation_setup}.

The main-text experiments use Prop-1 for MAE and Prop-3 for self-distilled DINOv3, selected to balance pretraining compute savings and downstream performance. Unless stated otherwise, we focus on Stacking for MAE and LIDAS for DINOv3 as representative configurations, with MIDAS also included in the DINOv3 layer-wise and attention comparisons. We first evaluate downstream performance and pretraining compute, including extended MAE training and distillation from a pretrained teacher (\cref{subsec:exp_downstream}), then examine intermediate representations (\cref{subsec:exp_layerwise_downstream}) and computational patterns (\cref{subsec:exp_computational}). The appendix complements these results with ablations of growing strategies, schedules, training configurations, and joint MAE encoder--decoder growth (\cref{app:ablation_study}), along with additional layer-wise and computational analyses (\cref{app:additional_layerwise_downstream,app:additional_computational_analyses}). It also examines robustness to missing visual evidence (\cref{app:additional_results_masking}), MAE performance with nonlinear probes and full fine-tuning (\cref{app:mae_adaptation}), and representation and weight similarity (\cref{app:representation_weight_similarity}).\looseness=-1

\subsection{Growing Improves Downstream Performance with Less Compute}
\label{subsec:exp_downstream}

To assess whether growing can improve representation quality while reducing pretraining cost, we compare grown models with their fixed-depth baselines on tasks covering both global and dense visual understanding. We evaluate frozen final-layer representations on ImageNet-1k \citep{deng2009imagenet} classification through $k$-NN and linear-probe top-1 accuracy, ADE20K \citep{zhou2019ade20k} segmentation through mean intersection over union (mIoU), and NYUv2 \citep{silberman2012nyu} depth estimation through root mean square error (RMSE). We additionally compute the total FLOPs of each model and report them relative to the matched fixed-depth baseline. For MAE, FLOPs calculation includes both the progressively grown encoder and the fixed decoder. For evaluations with trained heads, we report mean $\pm$ standard deviation across three downstream seeds for each model.

The results are shown in \cref{tab:main_downstream}. We observe that growing produces stronger MAE representations while substantially reducing pretraining compute. For ViT-L, the grown models use only $79.6\%$ of the baseline FLOPs, yet outperform the baseline across all downstream tasks with the appropriate growing strategy. Stacking gives the strongest overall gains, improving linear accuracy from $60.28$ to $63.64$ and mIoU from $26.29$ to $27.73$, while reducing RMSE from $0.6393$ to $0.5855$. LIDAS achieves the best $k$-NN accuracy. The same pattern holds at the larger ViT-H scale despite an even larger $24.8\%$ FLOP reduction, with LIDAS achieving the highest linear accuracy and mIoU, while Stacking achieves the highest $k$-NN accuracy and lowest RMSE. Stacking also improves all four metrics under WSD schedule (\cref{app:ablation_training_configuration}). Furthermore, the classification gains can extend to nonlinear probes and full fine-tuning (\cref{app:mae_adaptation}). Thus, growing is not merely a compute-saving approximation to fixed-depth training: with $20$--$25\%$ less pretraining compute, growing can yield better representations for both global and dense downstream tasks.

\begin{table}[t]
    \vspace{-0.5em}
    \centering
    \caption{
        \textbf{Downstream performance and pretraining compute of fixed-depth baseline and grown models across global and dense tasks.}
        Results are reported as mean $\pm$ standard deviation across three downstream seeds, while $k$-NN is deterministic.
        FLOPs are reported relative to the corresponding fixed-depth baseline. Across both MAE and DINOv3, growing improves downstream performance while reducing pretraining compute.
        }
    \label{tab:main_downstream}
    \small
    \setlength{\tabcolsep}{4pt}
    \renewcommand{\arraystretch}{0.9}
    \begin{tabular}{clccccc}
        \toprule
        & & Training & \multicolumn{2}{c}{ImageNet-1k}
        & ADE20K & NYU \\
        \cmidrule(lr){4-5}
        Model & Strategy
        & FLOPs (\%) $\downarrow$
        & $k$-NN $\uparrow$
        & Linear $\uparrow$
        & mIoU $\uparrow$
        & RMSE $\downarrow$ \\
        \midrule
        \multirow{4}{*}{%
        \rotatebox[origin=c]{90}{\shortstack{MAE\\ViT-L/16}}}
        & Baseline & $100.0$
        & $31.40$ & $60.28 \pm 0.07$
        & $26.29 \pm 0.03$ & $0.6393 \pm 0.0009$ \\
        \cmidrule(l){2-7}
        & Stacking & \multirow{3}{*}{$\mathbf{79.6}$}
        & $33.86$ & $\mathbf{63.64 \pm 0.08}$
        & $\mathbf{27.73 \pm 0.04}$ & $\mathbf{0.5855 \pm 0.0001}$ \\
        & MIDAS &
        & $32.21$ & $59.59 \pm 0.07$
        & $25.82 \pm 0.08$ & $0.6219 \pm 0.0004$ \\
        & LIDAS &
        & $\mathbf{34.09}$ & $61.08 \pm 0.08$
        & $26.54 \pm 0.08$ & $0.6292 \pm 0.0009$ \\
        \midrule
        \multirow{4}{*}{%
        \rotatebox[origin=c]{90}{\shortstack{MAE\\ViT-H/14}}}
        & Baseline & $100.0$
        & $36.59$ & $66.25 \pm 0.10$
        & $28.76 \pm 0.03$ & $0.5886 \pm 0.0006$ \\
        \cmidrule(l){2-7}
        & Stacking & \multirow{3}{*}{$\mathbf{75.2}$}
        & $\mathbf{37.99}$ & $66.05 \pm 0.05$
        & $29.66 \pm 0.03$ & $\mathbf{0.5680 \pm 0.0002}$ \\
        & MIDAS &
        & $33.83$ & $65.82 \pm 0.04$
        & $29.17 \pm 0.03$ & $0.5761 \pm 0.0002$ \\
        & LIDAS &
        & $34.60$ & $\mathbf{68.05 \pm 0.11}$
        & $\mathbf{30.18 \pm 0.02}$ & $0.5691 \pm 0.0009$ \\
        \midrule
        \multirow{3}{*}{%
        \rotatebox[origin=c]{90}{\shortstack{DINOv3\\ViT-L/16}}}
        & Baseline & $100.0$
        & $82.17$ & $83.40 \pm 0.04$
        & $44.70 \pm 0.10$ & $0.4650 \pm 0.0001$ \\
        \cmidrule(l){2-7}
        & MIDAS & \multirow{2}{*}{$\mathbf{86.1}$}
        & $\mathbf{82.23}$ & $83.71 \pm 0.06$
        & $45.64 \pm 0.10$ & $0.4614 \pm 0.0006$ \\
        & LIDAS &
        & $82.16$ & $\mathbf{83.73 \pm 0.04}$
        & $\mathbf{45.77 \pm 0.09}$ & $\mathbf{0.4523 \pm 0.0005}$ \\
        \bottomrule
    \end{tabular}
\end{table}

\begin{wrapfigure}[17]{r}{0.4\textwidth}
    \vspace{-0.8em}
    \centering
    \includegraphics[width=\linewidth]{
        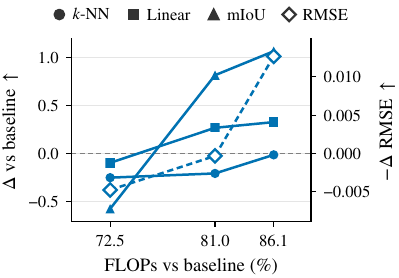
    }
    \captionsetup{font=footnotesize,skip=4pt}
    \caption{
    \textbf{DINOv3 compute--performance trade-off.} LIDAS gains over the fixed-depth baseline versus relative pretraining FLOPs for Prop-1--3 (left to right). Accuracy and mIoU gains use the left axis, and RMSE reduction uses the right axis. Allocating more training to deeper stages improves all four metrics while retaining compute savings.}
    \label{fig:flops_tradeoff_dinov3}
\end{wrapfigure}

The trend extends to the DINOv3 family. MIDAS and LIDAS use $86.1\%$ of baseline FLOPs while improving linear probing, segmentation, and depth estimation. LIDAS reaches $83.73$ linear accuracy, $45.77$ mIoU, and $0.4523$ RMSE versus $83.40$, $44.70$, and $0.4650$ for the fixed-depth baseline, with essentially unchanged $k$-NN accuracy. To further examine how these gains depend on the growing schedule, \cref{fig:flops_tradeoff_dinov3} compares LIDAS under Prop-1, Prop-2, and Prop-3 at the same total number of training steps. Allocating more steps to deeper stages progressively improves all metrics, while pretraining compute remains below the fixed-depth baseline under every schedule. Together with \cref{tab:main_downstream}, these results show that progressive growth reduces pretraining cost while improving downstream performance across global and dense prediction tasks. Full ablations of growing strategies and schedules are provided in \cref{app:ablation_growing_schedule_strategy}.

We additionally analyze how robust representations are when the visual information is removed from the input through random and semantic input masking  (\cref{app:additional_evaluations,app:additional_results_masking}). For MAE, Stacking improves classification robustness under both masking types, while for DINOv3, LIDAS improves classification and segmentation robustness under semantic masking. Thus, the downstream benefits of growing extend beyond clean inputs and persist even when visual evidence is missing.

\subsubsection{Extended MAE Training}
\label{subsec:exp_extended_mae_training}

\begin{figure}[t]
    \vspace{-1em}
    \centering  
    \includegraphics[width=0.98\linewidth]{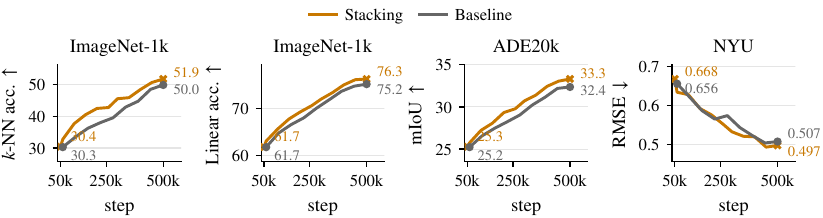}
    \caption{
        \textbf{Extended MAE training.} 
        Downstream performance of MAE trained with Stacking and its fixed-depth baseline.
        Stacking outperforms the baseline on all three tasks even after extended training.\looseness=-1
    }
    \label{fig:mae_extended_training}
\end{figure}

Our experiments so far train MAE variants for 50k steps. To examine whether the benefits of growing persist over the longer training duration according to the original MAE recipe, we train MAE ViT-L for 500k steps (approximately 1600 epochs). We compare Stacking with Prop-1 against a fixed-depth baseline. Both models use a learning-rate schedule defined over the full 500k training steps, while the growing schedule spans only the first 50k steps ($T_g = 50k$), after which the grown model continues training at full depth until 500k steps. As shown in \cref{fig:mae_extended_training}, Stacking outperforms the baseline across all three tasks at nearly all evaluated checkpoints, with its advantage persisting up to 500k steps. Final gains are approximately 1.9 and 1.1 percentage points in $k$-NN and linear accuracy, 0.9 points in mIoU, and a 0.01 reduction in RMSE. These results show that the benefits of growing persist after substantial additional training at full depth. Additional comparisons of training budgets and growth horizons are provided in \cref{app:ablation_training_growth_horizon}.

\subsubsection{Distillation from a Pretrained Teacher}
\label{subsec:exp_distillation_frozen_teacher}

\begin{table}[t]
    \centering
    \caption{
        \textbf{Distillation from a frozen pretrained teacher.}
        All grown models use Prop-1 schedule. Results are reported as mean $\pm$ standard deviation over three downstream seeds, while $k$-NN is deterministic. FLOPs are reported relative to the fixed-depth student. A frozen teacher runs at full depth throughout training, so growing saves less FLOPs than under self-distillation.
        }
    \label{tab:distillation}
    \small
    \setlength{\tabcolsep}{4pt}
    \renewcommand{\arraystretch}{0.9}
    \begin{tabular}{lccccc}
        \toprule
        & Training & \multicolumn{2}{c}{ImageNet-1k} & ADE20K & NYU \\
        \cmidrule(lr){3-4}
        Model & FLOPs (\%) $\downarrow$
        & $k$-NN $\uparrow$ & Linear $\uparrow$
        & mIoU $\uparrow$ & RMSE $\downarrow$ \\
        \midrule
        Frozen teacher & $100.0$ & $\mathbf{82.17}$ & $\mathbf{83.40 \pm 0.04}$ & $\mathbf{44.70 \pm 0.10}$ & $0.4650 \pm 0.0001$ \\
        \midrule
        \multicolumn{6}{l}{\textit{Frozen-teacher distillation}} \\
        \quad Fixed depth & $100.0$ & $81.73$ & $82.97 \pm 0.02$ & $43.43 \pm 0.10$ & $0.4674 \pm 0.0006$ \\
        \quad MIDAS & \multirow{2}{*}{$\mathbf{77.0}$} & $81.54$ & $82.82 \pm 0.03$ & $43.65 \pm 0.20$ & $\mathbf{0.4564 \pm 0.0007}$ \\
        \quad LIDAS & & $81.55$ & $82.92 \pm 0.03$ & $43.71 \pm 0.04$ & $\underline{0.4579 \pm 0.0010}$ \\
        \addlinespace[2pt]
        \multicolumn{6}{l}{\textit{Self-distillation}} \\
        \quad MIDAS & \multirow{2}{*}{$\mathbf{72.5}$} & $81.81$ & $83.15 \pm 0.03$ & $44.05 \pm 0.08$ & $0.4625 \pm 0.0005$ \\
        \quad LIDAS & & $\underline{81.92}$ & $\underline{83.31 \pm 0.03}$ & $\underline{44.12 \pm 0.08}$ & $0.4697 \pm 0.0007$ \\
        \bottomrule
    \end{tabular}
\end{table}

We next examine whether growing remains beneficial when distilling from a frozen pretrained teacher. We use our fixed-depth DINOv3 ViT-L trained on ImageNet-1k as a frozen teacher, and follow the original DINOv3 recipe for frozen-teacher distillation.\footnote{The public DINOv3 backbone checkpoints do not include the DINO and iBOT projection heads required for the distillation objective used here, motivating the use of our own pretrained teacher \citep{simeoni2025dinov3}.} We compare fixed-depth students with grown students using MIDAS and LIDAS under Prop-1 schedule. All students are trained for 125k steps with the same warmup-stable-decay (WSD) learning-rate schedule. The grown students therefore reach the same final architecture as the teacher. Full training details and student-recipe ablations are provided in \cref{app:distillation_setup,app:ablation_distillation_recipes}.

As shown in \cref{tab:distillation}, growing reduces FLOPs by $23\%$ relative to the fixed-depth student while retaining similar classification performance and improving dense prediction scores. MIDAS and LIDAS slightly reduce ImageNet $k$-NN accuracy by $0.19$ and $0.18$ percentage points, respectively, and linear accuracy by $0.15$ and $0.05$ points. In contrast, ADE20K mIoU increases from $43.43$ to $43.65$ and $43.71$, while NYUv2 RMSE decreases from $0.4674$ to $0.4564$ and $0.4579$. Both grown students also achieve lower depth estimation error than the teacher ($0.4650$), although the teacher retains stronger classification and segmentation performance. This task-specific advantage is reminiscent of Born-Again Networks, in which students with the same architecture can outperform their teachers \citep{furlanello2018born}.\looseness=-1

Compared with the Prop-1 self-distillation results in \cref{tab:distillation}, the grown frozen-teacher students achieve lower depth estimation error but weaker classification and segmentation scores. These comparisons also differ in learning-rate schedule (WSD versus cosine), so they do not isolate the effect of the teacher setting. Keeping the frozen teacher at full depth limits the relative FLOP savings to $23\%$, compared with $27.5\%$ under self-distillation. Thus, growing remains beneficial with a frozen teacher, reducing compute while maintaining similar classification performance and improving dense prediction.\looseness=-1

\subsection{Growing Improves Representations in Intermediate Layers}
\label{subsec:exp_layerwise_downstream}

To understand how growing affects the development of representations across depth, we investigate whether task-relevant information becomes accessible earlier in the network and whether this differs between global classification and dense prediction tasks. We train a separate task-specific head at each layer using the same protocols as the final-layer evaluations, alongside layer-wise $k$-NN classification. Extended layer-wise results are provided in \cref{app:additional_layerwise_downstream}.

\begin{figure}[t]
    \vspace{-1em}
    \centering
    \includegraphics[width=0.9\linewidth]{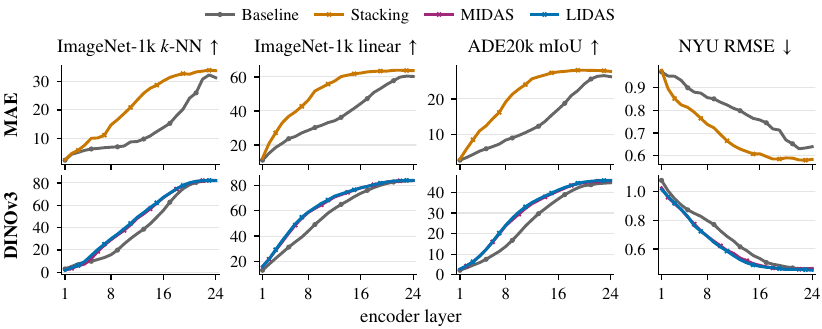}
    \caption{
        \textbf{Layer-wise downstream performance.}
        Frozen representations of MAE (top) and DINOv3 (bottom), with trained heads fitted separately at each layer. Growing improves performance across global and dense tasks, with larger gains in middle layers than at the final layer.
    }
    \label{fig:layerwise_downstream}
\end{figure}

As shown in \cref{fig:layerwise_downstream}, growing improves intermediate representations across both model families and all tasks, with particularly large advantages in the middle layers. For MAE, Stacking approaches its final classification and segmentation performance around layers 16--18 and plateaus in depth estimation, while the baseline continues to improve primarily in later layers. DINOv3 shows a similar shift, with MIDAS and LIDAS following closely aligned profiles and outperforming the baseline through much of the early and middle layers. Their margins narrow toward the final layers. We observe similar intermediate gains across strategies, model sizes, and teacher settings (\cref{app:additional_layerwise_downstream}). Thus, growing makes task-relevant information accessible at shallower depths across global and dense tasks, with intermediate improvements often substantially larger than final-layer gains.

\subsection{Growing Changes Computational Patterns}
\label{subsec:exp_computational}

The layer-wise evaluations show how useful the representations are at different depths. To complement this analysis, we study how growing changes the way models process information through attention analysis and layer-skipping interventions. We measure attention distance, spatial attention entropy, and query--key normalized mutual information (NMI), alongside patch-to-CLS and CLS-to-patch attention mass. Furthermore, we conduct layer-skipping interventions and bypass one transformer layer at a time and measure the relative change in the residual stream at subsequent layers compared with the unmodified model. The full setup is provided in \cref{app:computational_analyses}, with extended results in \cref{app:additional_computational_analyses}.\looseness=-1

\subsubsection{MAE: Repeated Cycles of Global and Local Attention}
\label{subsec:exp_computational_mae}

\begin{figure}[t]
    \vspace{-1em}
    \centering
    \includegraphics[width=0.9\linewidth]{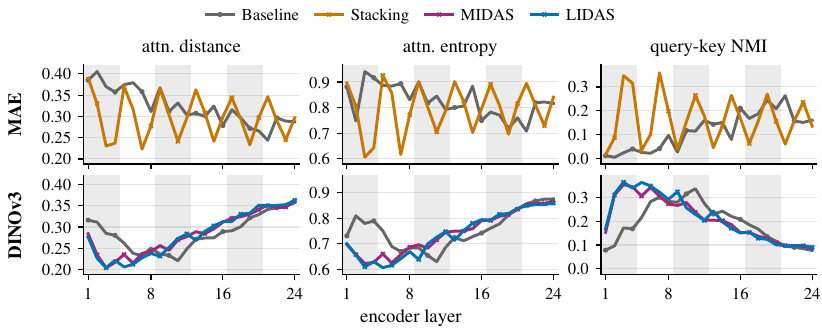}
    \caption{
        \textbf{Spatial attention patterns across depth.}
        Attention distance, spatial entropy, and query--key NMI for MAE (top) and DINOv3 (bottom). Alternating shaded bands mark groups of four layers. MAE Stacking exhibits a recurring four-layer pattern: distance and entropy peak near group starts and NMI peaks near third layers. DINOv3 growing shifts the phase of localized, query-dependent attention and the subsequent broadening of attention toward earlier layers.
    }
    \label{fig:attention_routing}
\end{figure}

We observe in \cref{fig:attention_routing} that Stacking produces a pronounced four-layer cycle in spatial attention, which matches the number of blocks inserted at each growing stage. 
Within successive four-layer groups, the first layer has relatively high attention distance and entropy and low query--key NMI, indicating global attention with little query dependence. Distance and entropy decrease in the second layer and reach their minima in the third, where NMI peaks, indicating localized, concentrated, and query-dependent attention. In the fourth layer, distance and entropy rise while NMI decreases, before broader attention resumes in the next group.
The cycle also changes across depth: distance peaks gradually become somewhat lower, and NMI peaks are lower in later groups. Thus, the groups share a recurring attention pattern while differing in its strength. The baseline shows layer-wise variation but lacks the same recurring pattern across all three metrics. These observations support a repeated sequence of global attention followed by localized, query-dependent attention within successive four-layer groups, and the recurring pattern holds for other growing strategies (see \cref{app:additional_computational_analyses}).

\begin{figure}[t]
    \vspace{-0.5em}
    \centering
    \includegraphics[width=0.9\linewidth]{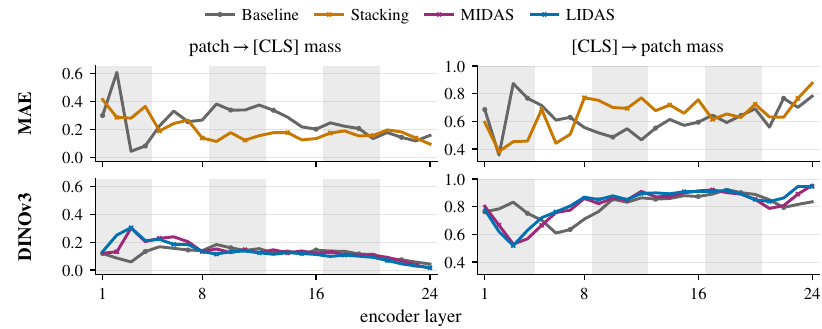}
    \caption{
        \textbf{Attention between CLS and patch tokens.}
        Patch-to-CLS (left) and CLS-to-patch mass (right) for MAE (top) and DINOv3 (bottom). Alternating shaded bands mark groups of four layers. MAE Stacking reduces patch-to-CLS attention and increases CLS-to-patch attention across much of the middle layers. 
        DINOv3 growing produces an earlier patch-to-CLS peak and CLS-to-patch recovery, with strongly asymmetric attention in the final layers.
    }
    \label{fig:attention_cls}
\end{figure}

The CLS-attention profiles in \cref{fig:attention_cls} show how growing changes attention between CLS and patch tokens. Across layers 8--15, Stacking directs less attention from patches to CLS and more attention from CLS to patches than the baseline. The repeating pattern is weaker and less regular than in the spatial metrics in \cref{fig:attention_routing}, suggesting that the four-layer cycle is mainly a feature of spatial attention rather than all token interactions.

The layer-skipping results in \cref{fig:layer_skip} furthermore show how Stacking changes layer influence across depth. Skipping the earliest layers produces substantially larger changes than in the baseline, and the strongest early effect persists through the final layers. Skipping several middle layers also has stronger effects on later representations. In contrast, skipping later layers generally changes subsequent residual streams less than in the baseline. This reduced late-layer sensitivity is consistent with the earlier saturation of MAE's downstream readouts observed in \cref{fig:layerwise_downstream} and suggests a computational shift towards earlier layers. When training for 500k steps, the four-layer cycle and stronger effects of skipping early layers persist even as layer-wise readouts become more similar and weight alignment weakens (\cref{app:additional_computational_analyses,app:representation_weight_similarity}). Together, these results identify recurring attention regimes alongside a shift toward stronger early-layer influence and reduced late-layer sensitivity.\looseness=-1

\subsubsection{DINOv3: Earlier Transition from Local to Global Attention}
\label{subsec:exp_computational_dino}

\begin{figure}[t]
    \vspace{-1em}
    \centering
    \includegraphics[width=\linewidth]{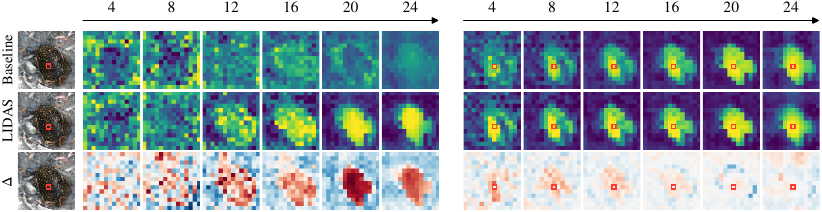}
    \caption{
    \textbf{Spatial feature similarities across depth for DINOv3.} 
    Rows show the fixed-depth baseline, LIDAS, and delta between the baseline and LIDAS, and columns show encoder layers for CLS--patch (left) and patch-to-patch (right) cosine similarities. Red squares mark the anchor patch. LIDAS shows clearer similarity patterns over object regions at earlier layers.
    }
    \label{fig:qualitative_dinov3}
\end{figure}

Self-distilled DINOv3 shows a different pattern. \cref{fig:attention_routing} shows a shift in the timing of attention regimes rather than MAE's pronounced four-layer cycle. 
Both MIDAS and LIDAS exhibit low attention distance and entropy and high NMI around layers 3--6. The baseline develops this combination later, with its shortest distance and a prominent NMI peak around layer 11. Growing therefore produces the phase of localized, concentrated, and query-dependent attention earlier in the network.
Attention distance and entropy then increase across much of the remaining network, while NMI generally decreases. As a result, the grown models show broader and less query-dependent spatial attention across much of the middle layers, where the baseline is still in its more localized phase. The profiles become more similar toward the final layers.

The CLS-attention profiles in \cref{fig:attention_cls} provide complementary evidence of earlier changes in token interactions. Both MIDAS and LIDAS show a patch-to-CLS peak around layer 3, where the baseline remains low before developing a smaller peak later. Meanwhile, CLS-to-patch mass drops significantly at layer 3 and then rises by layer 8, recovering earlier than in the baseline. Throughout the middle layers, patches assign similar or less attention to CLS than in the baseline, while CLS consistently assigns more attention to patches. By the final layer, CLS attends predominantly to patches, while patches allocate little attention to CLS. This suggests a shift towards earlier computation as observed in MAE.\looseness=-1

The layer-skipping results in \cref{fig:layer_skip} show smaller differences between LIDAS and its baseline than between MAE Stacking and its baseline. Nevertheless, the difference map shows the same pattern: Skipping early layers has stronger effects than the baseline in subsequent middle-layer readouts, while skipping late layers has a smaller effect, particularly in nearby readouts.

To complement the quantitative analyses, \cref{fig:qualitative_dinov3} compares CLS--patch and patch--patch cosine similarities in LIDAS and the fixed-depth baseline. LIDAS develops clearer CLS--patch patterns over object regions at layers 12--16, while the baseline's patterns are less coherent. Patch--patch similarities also form a coherent region over the object earlier. The visualization protocol and additional examples are provided in \cref{app:computational_analyses,app:qualitative_feature_maps}. Taken together, the attention and intervention results indicate an earlier transition from local to global attention and a shift in layer influence, while the qualitative maps illustrate changes in spatial feature organization across depth. These findings complement the improved intermediate representations observed in the downstream evaluations (\cref{subsec:exp_layerwise_downstream}).

\section{Discussion and Conclusion}
\label{sec:conclusion}

We study progressive depth growth as an efficient pretraining strategy for self-supervised vision transformers. Across MAE and DINOv3, suitable growth configurations improve classification, segmentation, and depth estimation while reducing pretraining compute. These benefits extend to intermediate representations, where improvements often exceed final-layer gains. Growing also changes internal processing: MAE develops recurring attention patterns, while DINOv3 exhibits earlier local-to-global attention progression. Layer-skipping interventions indicate stronger early-layer influence and reduced sensitivity to many later-layer interventions in both families. MAE's final-layer gains and recurring patterns persist after prolonged full-depth training. Together, these findings establish simple block copying as a practical pretraining approach that influences both representation quality and computational organization. 

Beyond these gains in pretraining efficiency and final-layer performance, our findings on intermediate representations and internal computations suggest several potential applications.
Stronger intermediate representations could support early-exit inference \citep{xu2023lgvit} and cheaper feature extraction for large-scale data curation \citep{simeoni2025dinov3}. They also motivate exploring early-exit teachers in self-distillation, using intermediate representations to generate less expensive training targets. Stronger intermediate spatial features could also support adaptive dense prediction, which allows confidently predicted regions to exit early while harder regions receive further processing \citep{tang2023dynamic}. Extending growing to autoregressive visual or multimodal models could additionally support self-speculative decoding, with shallow layers drafting tokens for deeper layers to verify \citep{elhoushi2024layerskip}. These possibilities extend the potential benefits of growing from efficient pretraining to more efficient use of learned representations.

\subsection*{AI use statement}

In this work, we used generative AI tools, including coding agents, to assist with method implementation and interpretation of experimental results. We have not used generative AI tools to develop theoretical models or conceptual frameworks, formulate mathematical claims, propose or refine hypotheses, design or provide feedback on research methodology or experiments, translate content, clean or reformat datasets, or conduct qualitative or thematic data analysis, and tasks involving synthetic dataset generation or mathematical proofs are not applicable to this work. Additionally, we used generative AI tools for literature retrieval and discovery, including finding related work, for drafting and polishing the manuscript, and for developing code to plot experimental results. We have reviewed all AI-assisted work, including text, code, and figures. We take responsibility for the final content of this work, including all text, claims, and artifacts produced with the aid of generative AI.

\subsection*{Reproducibility statement}
The information needed to reproduce our experiments is provided in the main text and appendix. Model architectures, pretraining configurations, and growth procedures and schedules are described in \cref{sec:experiments,app:training_setup}. Downstream evaluation protocols, data preprocessing, hyperparameters, and procedures for the layer-wise and computational analyses are detailed in \cref{app:evaluation_setup}. We build on publicly available implementations of MAE and DINOv3, which are properly cited in the paper, and use publicly available datasets.

\section*{Acknowledgments}
The authors would like to thank Max Seitzer for insightful discussions and support throughout this work.
This work was partially supported by the Wallenberg AI, Autonomous Systems and Software Program (WASP), funded by the Knut and Alice Wallenberg Foundation, and by the Helmholtz Foundation Model Initiative, supported by the Helmholtz Association.
The authors thank the Gauss Centre for Supercomputing e.V. (GCS) for providing computing time on the Supercomputer JUPITER at the Jülich Supercomputing Centre via the John von Neumann Institute for Computing. JUPITER is supported by the EuroHPC JU and GCS through funding by the European Commission, the German Federal Ministry of Research, Technology and Space, and the Ministry of Culture and Science of the State of North Rhine-Westphalia. The computations were also enabled by the Berzelius resource, provided by the Knut and Alice Wallenberg Foundation at the National Supercomputer Centre. Furthermore, the authors appreciate the computational resources provided by the National High Performance Computing Centre (www.nhr.kit.edu). The research presented is supported by the TUM Georg Nemetschek Institute Artificial
Intelligence for the Built World.

\bibliography{references}
\bibliographystyle{iclr2027_conference}

\clearpage
\appendix

\section{Training and Growing Setup}
\label{app:training_setup}

Here we describe the pretraining recipes, growth implementation, and frozen-teacher distillation setting used in our experiments. We build on the original MAE \citep{he2022masked} and DINOv3 \citep{simeoni2025dinov3} implementations, using the training configurations reported below. Unless explicitly stated, remaining implementation details follow the corresponding publicly available code. Unless stated otherwise, comparisons use the same number of steps and batch size, and grown models reach the architecture of their fixed-depth counterparts. Training budgets therefore match the number of images seen by baselines and grown models, while the intermediate model depths and pretraining compute differ. Downstream evaluation protocols are described in \cref{app:evaluation_setup}.

\subsection{Training Setup}

\textbf{Data and implementation.}~
All models are trained on the ImageNet-1k \citep{deng2009imagenet} training split which contains 1,281,167 images. Both families use 32 NVIDIA GH200 GPUs for training. We use the default ImageNet-1k input normalization with RGB mean of $(0.485,0.456,0.406)$ and standard deviation of $(0.229,0.224,0.225)$.

\subsubsection{MAE}

\textbf{Architecture and objective.}~
We use ViT-L/16 and ViT-H/14 encoders with the configurations in \cref{tab:app_mae_config} and fixed positional embeddings. Following the original MAE recipe, we randomly mask 75\% of image patches and pass only the visible patches and CLS token through the encoder. The decoder reconstructs per-patch normalized pixels from the encoded visible patches and mask tokens, using mean squared error on masked patches only. The 8-layer decoder remains fixed except in the joint-growth experiment in \cref{app:growing_setup}.

\textbf{Augmentation and optimization.}~
We use random resized cropping to $224\times224$ with scale range $[0.2,1.0]$ and bicubic interpolation, followed by horizontal flipping and ImageNet normalization. We use AdamW with constant weight decay and no layer-wise learning-rate decay. The principal ViT-L comparison uses linear warmup followed by cosine decay to zero. ViT-H uses warmup--stable--decay (WSD): linear warmup, a constant learning rate, and linear decay to zero over the final 20\% of training. \cref{tab:app_mae_budgets} lists the effective peak learning rates, warmup lengths, and additional ViT-L WSD settings.

\begin{table}[t]
    \centering
    \caption{Training hyperparameters of MAE. Encoder depths refer to the final architecture. The decoder configuration is shared by both model sizes. Learning rates are the effective peak values used by the optimizer.}
    \label{tab:app_mae_config}
    \small
    \renewcommand{\arraystretch}{1.12}
    \begin{tabularx}{\linewidth}{@{}lXX@{}}
        \toprule
        Configuration & ViT-L/16 & ViT-H/14 \\
        \midrule
        Input resolution & $224\times224$ & $224\times224$ \\
        Patch size & $16\times16$ & $14\times14$ \\
        Encoder depth / width & 24 / 1024 & 32 / 1280 \\
        Attention heads / MLP ratio & 16 / 4 & 16 / 4 \\
        Decoder depth / width & 8 / 512 & 8 / 512 \\
        Decoder attention heads & 16 & 16 \\
        Masking ratio & 0.75 & 0.75 \\
        Reconstruction target & Per-patch normalized pixels & Per-patch normalized pixels \\
        Batch size & 4096 & 4096 \\
        Gradient accumulation & 1 & 1 \\
        Optimizer & AdamW & AdamW \\
        Adam $(\beta_1,\beta_2)$ & $(0.9,0.95)$ & $(0.9,0.95)$ \\
        Weight decay & 0.05 & 0.05 \\
        Gradient clipping (norm) & 1.0 & 1.0 \\
        Stochastic depth & 0 & 0 \\
        Main training steps & 50,000 & 50,000 \\
        Main LR schedule & Cosine & WSD, linear cooldown \\
        Main peak LR & $2.4\times10^{-3}$ & $1.0\times10^{-3}$ \\
        Main LR warmup steps & 1,250 & 1,250 \\
        Final LR & 0 & 0 \\
        \bottomrule
    \end{tabularx}
\end{table}

\begin{table}[t]
    \centering
    \caption{Training budgets and schedule variants for MAE.}
    \label{tab:app_mae_budgets}
    \small
    \setlength{\tabcolsep}{5pt}
    \renewcommand{\arraystretch}{1.12}
    \begin{tabular}{@{}llrrrr@{}}
        \toprule
        Model & LR schedule & Steps & Peak LR & Warmup & Growth horizon \\
        \midrule
        ViT-L & Cosine & 50,000 & $2.4\!\times\!10^{-3}$ & 1,250 & 50,000 \\
        ViT-L & Cosine & 500,000 & $2.4\!\times\!10^{-3}$ & 12,500 & 50,000 \\
        ViT-L & WSD & 50,000 & $1.0\!\times\!10^{-3}$ & 1,250 & 50,000 \\
        ViT-L & WSD & 500,000 & $1.0\!\times\!10^{-3}$ & 12,500 & 50,000 \\
        ViT-H & WSD & 50,000 & $1.0\!\times\!10^{-3}$ & 1,250 & 50,000 \\
        ViT-H & WSD & 500,000 & $1.0\!\times\!10^{-3}$ & 12,500 & 50,000 \\
        \bottomrule
    \end{tabular}
\end{table}

\textbf{Extended MAE training.}~
The standard 50k-step budget corresponds to approximately 160 epochs. For extended training, we follow the original MAE recipe and train MAE ViT-L for 500k steps (approximately 1,600 epochs), with the learning-rate schedule defined over the full 500k steps from the outset and a 12,500-step warmup. Growth follows Prop-1 over a separate 50k-step horizon: the encoder reaches depth 24 at step 35,710 and continues at full depth until step 500,000. No expansion or schedule restart occurs at step 50,000. The main extended comparison uses Stacking with cosine schedule. Additional WSD runs include all three strategies. Extended ViT-H training follows the same 500k-step procedure, using the ViT-H WSD recipe with a 12,500-step warmup and a 50k-step Prop-1 growth horizon. The grown encoder reaches its full depth of 32 layers at step 38,885 and remains at full depth for the rest of training.

\subsubsection{DINOv3}

\textbf{Architecture and objective.}~
Following the public recipe for ImageNet training of DINOv3, we use ViT-L/16 with 24 layers, width 1024, 16 attention heads, MLP ratio 4, and LayerScale initialized to $10^{-5}$. The backbone uses rotary positional embeddings (RoPE) with separate coordinate normalization along each spatial axis and no coordinate augmentation. Register tokens and Gram anchoring are disabled. Our ImageNet-1k recipe uses the exact suggested recipe, with DINO image-level objective, the iBOT patch-level objective, and KoLeo regularization:

\begin{equation}
    \mathcal{L}_{\mathrm{SSL}}
    = \mathcal{L}_{\mathrm{DINO}}
    + \mathcal{L}_{\mathrm{iBOT}}
    + 0.1\,\mathcal{L}_{\mathrm{KoLeo}}.
    \label{eq:app_dino_loss}
\end{equation}

\textbf{Views, augmentation, and targets.}~
Each image produces two global crops of $224\times224$ pixels with scale range $[0.32,1.0]$ and eight local crops of $96\times96$ pixels with scale range $[0.05,0.32]$. We use the reference multi-crop photometric pipeline, including color jitter, grayscale conversion, Gaussian blur, and solarization, together with horizontal flipping and ImageNet normalization. The teacher processes global crops, and the student processes global and local crops. The DINO objective excludes same-view global matches. DINO and iBOT use separate three-layer projection heads with hidden dimension 2048, bottleneck dimension 256, and 65,536 prototypes. Both student temperatures are 0.1, and teacher targets use three Sinkhorn--Knopp iterations. KoLeo is applied to global CLS token features using a non-distributed implementation. For iBOT, half the global crops receive spatial block masks with random-patch completion. Their ratios follow a shuffled ladder spanning the configured range $[0.1,0.5]$. Local crops are unmasked, and masking retains the backbone's sequence length.

\begin{table}[t]
    \centering
    \caption{Training hyperparameters of DINOv3. All columns use ViT-L/16, 125,000 training steps, batch size of 2048, and the same crop and head configurations. Peak learning rates are reported before batch-size scaling and parameter-group multipliers. The WSD Self-distilled column describes the grown reference which uses $1-\sqrt{s}$ cooldown. The fixed-depth schedule ablation in \cref{tab:app_dinov3_knobs} instead uses linear WSD as its reference.}
    \label{tab:app_dino_config}
    \small
    \setlength{\tabcolsep}{4pt}
    \renewcommand{\arraystretch}{1.12}
    \begin{tabularx}{\linewidth}{@{}lXXX@{}}
        \toprule
        Configuration & Self-distilled, cosine & Self-distilled, WSD & Frozen-teacher, WSD \\
        \midrule
        Optimizer & AdamW & AdamW & AdamW \\
        Adam $(\beta_1,\beta_2)$ & $(0.9,0.999)$ & $(0.9,0.999)$ & $(0.9,0.999)$ \\
        Batch size & 2048 & 2048 & 2048 \\
        Peak LR & $10^{-3}$ & $5\times10^{-4}$ & $5\times10^{-4}$ \\ \\
        Final LR & $10^{-6}$ & $10^{-6}$ & $10^{-6}$ \\
        LR warmup steps & 12,500 & 12,500 & 12,500 \\
        LR decay & Cosine after warmup & $1-\sqrt{s}$, final 25k & $1-\sqrt{s}$, final 25k \\
        Weight decay & $0.04\to0.4$ & $0.04\to0.4$ & $0.04\to0.4$ \\
        Layer-wise LR decay & 0.9 & 0.9 & 0.99 \\
        Patch-embedding LR multiplier & 0.2 & 0.2 & 0.2 \\
        Gradient clipping (norm) & 3.0 & 3.0 & 3.0 \\
        Uniform stochastic-depth rate & 0.3 & 0.1 & 0 \\
        Head last-layer freeze (steps) & 1,250 & 1,250 & 0 \\
        Target-teacher temperature & $0.04\to0.07$ & $0.04\to0.07$ & 0.07 \\
        Temperature warmup (steps) & 37,500 & 37,500 & None \\
        Student EMA momentum & $0.992\to1$ & $0.992\to1$ & $0.992\to1$ \\
        Target teacher & Student EMA & Student EMA & Frozen model \\
        Student global-crop masking & Enabled & Enabled & Disabled \\
        \bottomrule
    \end{tabularx}
\end{table}

\textbf{Optimization and teacher updates.}~
Following DINOv3's public training recipe, training uses 125k steps at global batch size 2048, approximately 200 ImageNet epochs. \cref{tab:app_dino_config} reports peak learning rates before batch-size scaling. Layer-wise decay and the patch-embedding multiplier further scale parameter-group learning rates. Cosine training uses linear warmup followed by cosine decay. For the grown WSD reference, the learning rate remains at its peak after warmup until step 100k, then decays over the final 25k steps according to $\eta(s)=\eta_{\min}+(\eta_{\max}-\eta_{\min})(1-\sqrt{s})$, where $s\in[0,1]$ is cooldown progress. Weight decay and student-EMA momentum follow cosine schedules over the full training budget. In ordinary self-distillation, this EMA supplies the teacher targets, with temperature increasing linearly during the warmup (\cref{tab:app_dino_config}). Stochastic depth is uniform across layers, and inserted layers retain the same per-layer rate after growth.

\subsubsection{Training hyperparameter selection}
For the main MAE ViT-L comparisons, we retain the cosine learning-rate schedule and default base learning rate of $1.5\times10^{-4}$ from the original MAE recipe \citep{he2022masked}. At global batch size 4096, this gives an effective peak learning rate of $2.4\times10^{-3}$. The fixed-depth and grown models use the same recipe. The learning-rate and schedule sweeps in \cref{app:ablation_training_configuration} assess sensitivity to these choices. For MAE ViT-H, we use a WSD learning-rate schedule with an
effective peak learning rate of $10^{-3}$, adopting the
best-performing WSD learning rate from the MAE ViT-L sweep
in \cref{app:ablation_training_configuration}.

For DINOv3, sweeps on fixed-depth baseline inform the adopted learning rates and optimization settings, using final-layer ImageNet-1k linear-probe accuracy as the primary criterion and the other downstream metrics to assess trade-offs across tasks. The configurations and local references are specified in \cref{app:ablation_training_configuration}. Selection uses the same ImageNet-1k validation split used for reporting, without a separate held-out selection split.

\subsection{Growing Setup}
\label{app:growing_setup}

Standard growth starts from four Transformer layers and adds four layers per stage. ViT-L models use six stages with depths $(4,8,12,16,20,24)$, while MAE ViT-H uses eight stages with depths $(4,8,12,16,20,24,28,32)$. Everything else such as width, attention-head count, patch size, and non-block architecture remain fixed. MAE growth acts on the encoder only by default, with its 8-layer decoder unchanged throughout training. The main MAE comparisons use Stacking, MIDAS, and LIDAS with Prop-1 schedule, and the main DINOv3 comparisons use MIDAS and LIDAS with Prop-3 schedule. We exclude Stacking for DINOv3 because of its poor performance with Prop-1 schedule. We report full ablations on growing schedule and strategy in \cref{app:ablation_growing_schedule_strategy}.

\begin{figure}
    \centering
    \includegraphics[width=\linewidth]{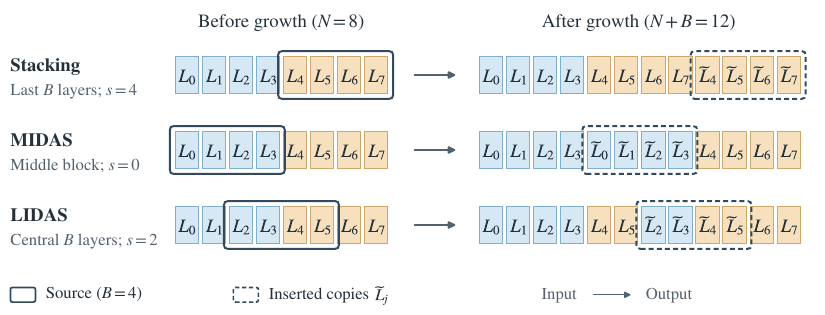}
    \caption{
    A simple illustration of growing strategies.
    Each row illustrates one growing stage with $N=8$ and $B=4$.
    Solid outlines mark the source layers and dashed outlines mark their
    copies $\widetilde{L}_j$, inserted immediately after the source.
    Stacking copies the final $B$ layers, MIDAS the middle $B$-layer
    block, and LIDAS the central $B$-layer slice.
    Colors identify the two original four-layer blocks.}
    \label{fig:growing_strategies}
\end{figure}

\textbf{Layer-copy mappings.}~
Before growth, the model has ordered layers $(L_0,\ldots,L_{N-1})$. Each event inserts $B$ layers. Each strategy selects a contiguous source slice beginning at index $s$ and inserts its copy immediately after that slice:

\begin{equation}
    (L_0,\ldots,L_{s+B-1},\;
    \widetilde{L}_{s},\ldots,\widetilde{L}_{s+B-1},\;
    L_{s+B},\ldots,L_{N-1}),
    \label{eq:app_layer_copy}
\end{equation}

where $\widetilde{L}_j$ initially has exactly the parameters of $L_j$, stored independently. The source index is

\begin{equation}
    s=\begin{cases}
        N-B, & \text{Stacking},\\
        B\left\lfloor (N/B-1)/2\right\rfloor, & \text{MIDAS},\\
        \left\lceil (N-B)/2\right\rceil, & \text{LIDAS}.
    \end{cases}
\end{equation}

Stacking copies the final $B$ layers and appends them to the network. MIDAS selects the middle $B$-layer chunk, choosing the earlier chunk when there are two central chunks. LIDAS selects the central $B$-layer slice without restricting it to chunk boundaries. \cref{tab:app_copy_slices} gives the exact source slices. Copies occupy target indices $s+B$ through $s+2B-1$, and subsequent original layers shift by $B$. Copied layers receive no noise, interpolation, or residual rescaling. Parameters outside the growing blocks are retained. All strategies have the same mapping at the first $4\to8$ transition, and LIDAS and MIDAS differ at two of the five ViT-L growth events.

\begin{table}[t]
    \centering
    \caption{Source-layer indices copied at each standard growth event ($B=4$). Indices are zero-based and ranges are inclusive. Each slice is inserted immediately after its source, as in \cref{eq:app_layer_copy}. The final two transitions apply only to ViT-H.}
    \label{tab:app_copy_slices}
    \small
    \renewcommand{\arraystretch}{1.12}
    \begin{tabular}{@{}lccc@{}}
        \toprule
        Depth transition & Stacking & MIDAS & LIDAS \\
        \midrule
        $4\to8$ & 0--3 & 0--3 & 0--3 \\
        $8\to12$ & 4--7 & 0--3 & 2--5 \\
        $12\to16$ & 8--11 & 4--7 & 4--7 \\
        $16\to20$ & 12--15 & 4--7 & 6--9 \\
        $20\to24$ & 16--19 & 8--11 & 8--11 \\
        $24\to28$ & 20--23 & 8--11 & 10--13 \\
        $28\to32$ & 24--27 & 12--15 & 12--15 \\
    \bottomrule
    \end{tabular}
\end{table}

\textbf{Stage durations.}~
For $K$ stages and growth-schedule horizon $T_g$, Prop-$\alpha$ assigns stage $i$ a duration proportional to $i^\alpha$. We compute integer stage lengths as
\begin{equation}
    \Delta_i=\left\lfloor T_g\frac{i^\alpha}{\sum_{j=1}^{K}j^\alpha}\right\rfloor
    \quad(i<K),\qquad
    \Delta_K=T_g-\sum_{i=1}^{K-1}\Delta_i.
    \label{eq:app_stage_duration}
\end{equation}

Thus, all rounding remainder is assigned to the final stage. Stage $i$ occupies the half-open step interval $[t_{i-1},t_i)$, where $t_0=0$ and $t_i=\sum_{j=1}^{i}\Delta_j$; growth occurs at $t_1,\ldots,t_{K-1}$. \cref{tab:app_stage_boundaries} reports the exact boundaries for the main configurations used in the paper. For standard DINOv3 runs, $T_g=125$k includes the final 25k steps even under WSD. The 500k-step MAE runs instead use $T_g=50$k and continue at full depth for the remaining training steps.

\begin{table}[t]
    \centering
    \caption{Cumulative stage endpoints in optimizer steps. Columns correspond to the depth used during the stage ending at that step. The last endpoint is the growth-schedule horizon, not another growth event. The 500k-step MAE runs reuse the 50k Prop-1 endpoints and then continue to step 500,000 at their final depth.}
    \label{tab:app_stage_boundaries}
    \small
    \setlength{\tabcolsep}{4pt}
    \renewcommand{\arraystretch}{1.12}
    \begin{tabular}{@{}llrrrrrr@{}}
        \toprule
        Model & Schedule & Depth 4 & 8 & 12 & 16 & 20 & 24 \\
        \midrule
        MAE ViT-L & Prop-1 & 2,380 & 7,141 & 14,283 & 23,806 & 35,710 & 50,000 \\
        MAE ViT-L & Prop-2 & 549 & 2,746 & 7,691 & 16,482 & 30,218 & 50,000 \\
        MAE ViT-L & Prop-3 & 113 & 1,020 & 4,081 & 11,337 & 25,509 & 50,000 \\
        DINOv3 ViT-L & Prop-1 & 5,952 & 17,856 & 35,713 & 59,522 & 89,283 & 125,000 \\
        DINOv3 ViT-L & Prop-2 & 1,373 & 6,867 & 19,229 & 41,207 & 75,547 & 125,000 \\
        DINOv3 ViT-L & Prop-3 & 283 & 2,550 & 10,203 & 28,343 & 63,773 & 125,000 \\
        \bottomrule
        \end{tabular}
        \par\medskip
        \begin{tabular}{@{}lrrrrrrrr@{}}
        \toprule
        MAE ViT-H, Prop-1 & Depth 4 & 8 & 12 & 16 & 20 & 24 & 28 & 32 \\
        \midrule
        Stage endpoint & 1,388 & 4,165 & 8,331 & 13,886 & 20,830 & 29,163 & 38,885 & 50,000 \\
        \bottomrule
    \end{tabular}
\end{table}

\textbf{Optimizer state and schedule continuity.}~
At each transition, we expand a saved checkpoint and resume training the larger model. Each stage starts a new data loader, and each inserted parameter inherits independent copies of its source parameter's AdamW first and second moments and optimizer step counter. Existing parameters retain their optimizer state. The global iteration counter is preserved, so global-iteration hyperparameters such as learning rate, weight decay, teacher momentum, and teacher temperature continue at the same position in their global schedules. We do not restart warmup at growth transitions. MAE retains optimizer-group hyperparameters, and DINOv3 rebuilds parameter groups for the expanded architecture, recomputing layer-wise learning-rate multipliers at the new depth. Consequently, individual DINOv3 parameter learning rates may change at a transition even though the global learning-rate schedule continues.

\textbf{DINOv3 teacher settings.}~
In ordinary DINOv3 self-distillation, we copy the expanded student backbone into the teacher at each growth event. The teacher retains its EMA DINO and iBOT heads. The EMA module is synchronized with this teacher state, and EMA updates then resume using the current global momentum. Thus, the teacher backbone is reset at transitions rather than expanded from its own previous EMA blocks. The frozen target teacher used for pretrained-teacher distillation is handled separately, as described in \cref{app:distillation_setup}.

\textbf{Joint MAE encoder--decoder growth.}~
To examine decoder growth, we retain the ViT-L encoder's Stacking configuration, with four initial layers and four-layer increments under Prop-1 over 50k steps, and replace the fixed decoder with a growing decoder. The decoder starts with two layers and grows to eight using two-layer LIDAS insertions and its own four-stage Prop-1 schedule. Its expansion steps are 5,000, 15,000, and 30,000. For decoder depths 2, 4, and 6, the copied source slices are respectively $(0,1)$, $(1,2)$, and $(2,3)$, each inserted immediately after its source. Encoder and decoder growth events are merged on the same global step axis, giving the nine segments shown in \cref{tab:app_joint_growth}. The encoder schedule, optimizer, loss, batch size, and cosine learning-rate schedule match the encoder-only Stacking run, and only the decoder depth progression changes. Both configurations end with the same 24-layer encoder and 8-layer decoder.

\begin{table}[t]
    \centering
    \caption{Joint MAE growth with Stacking for the encoder and LIDAS for the decoder.}
    \label{tab:app_joint_growth}
    \small
    \renewcommand{\arraystretch}{1.12}
    \begin{tabular}{@{}lcc@{}}
        \toprule
        Step interval & Encoder depth & Decoder depth \\
        \midrule
        $[0,2380)$ & 4 & 2 \\
        $[2380,5000)$ & 8 & 2 \\
        $[5000,7141)$ & 8 & 4 \\
        $[7141,14283)$ & 12 & 4 \\
        $[14283,15000)$ & 16 & 4 \\
        $[15000,23806)$ & 16 & 6 \\
        $[23806,30000)$ & 20 & 6 \\
        $[30000,35710)$ & 20 & 8 \\
        $[35710,50000)$ & 24 & 8 \\
        \bottomrule
    \end{tabular}
\end{table}

\subsection{Distillation Setup}
\label{app:distillation_setup}

\textbf{Pretrained teacher.}~
We use the final EMA teacher checkpoint of our own fixed-depth DINOv3 ViT-L baseline, trained from scratch on ImageNet-1k for 125k steps with the cosine recipe in \cref{tab:app_dino_config}. The checkpoint contains the backbone and both DINO and iBOT heads. The teacher is excluded from gradient and EMA updates during student training and is kept fixed throughout training.

\textbf{Student training.}~
Fixed-depth students start with 24 layers, whereas grown students start with four layers and follow the six-stage Prop-1 schedule over 125k steps using LIDAS or MIDAS. Students are randomly initialized and do not inherit pretrained teacher weights. We retain the objective in \cref{eq:app_dino_loss}, but obtain targets from the frozen pretrained teacher instead of the student's EMA. Teacher temperature is fixed at 0.07 from the first step. Distillation uses WSD with a peak learning rate of $5\times10^{-4}$, 12,500 warmup steps, and $1-\sqrt{s}$ cooldown over the final 25k steps. Stochastic depth is disabled, layer-wise learning-rate decay is 0.99, and the projection heads' final layers are trainable from the start. Remaining hyperparameters are given in \cref{tab:app_dino_config}.

\textbf{Patch targets and student EMA.}~
During frozen-teacher distillation, the student's global crops are unmasked. Patch-selection indices are still sampled for the iBOT loss, and the frozen teacher's iBOT head supplies targets at those selected positions. A separate EMA of the student is maintained for evaluation, using the scheduled momentum from 0.992 to 1.0. The exported evaluation checkpoint contains this student EMA, not the frozen pretrained teacher. The target teacher remains fixed as the student grows.
For grown models, at each growth transition, the student's EMA backbone and both projection heads are reset to the expanded student's weights. Previous EMA history is discarded, and averaging resumes at the current global momentum without restarting its schedule. For Prop-1, the final EMA is therefore accumulated after the last expansion at step 89,283, over the remaining 35,717 updates. The fixed-depth student's EMA does not undergo these growth-triggered resets.

\section{Evaluation Setup}
\label{app:evaluation_setup}

Here we describe the evaluation setup used in our study. We evaluate classification, semantic segmentation, and monocular depth estimation using the evaluation pipeline of the DINOv3 repository \citep{simeoni2025dinov3}, adapted to load our MAE and DINOv3 checkpoints and read out intermediate layers. Routine preprocessing, loss implementations, and inference operations follow the corresponding DINOv3 evaluation setup unless stated otherwise. We also describe the attention analyses, interventions, and additional evaluations used to characterize the learned representations.

\subsection{Downstream Evaluation}
\label{app:downstream_evaluation}

\textbf{Backbone features.}~
MAE and DINOv3 backbones are frozen and run in evaluation mode, except in the explicitly identified MAE fine-tuning experiment explained in \cref{app:additional_evaluations}. The main performance tables use the final-layer representation. For layer-wise evaluation, we retain the first $\ell$ blocks and apply the same feature-extraction and evaluation protocol to their output. Each layer receives its own trained task head. \cref{tab:app_eval_features} summarizes the feature choices for each downstream task.

\begin{table}[t]
    \centering
    \caption{Feature extraction for frozen-backbone evaluation. $D$ is the backbone width. The backbone's output LayerNorm is applied at each evaluated layer. Dense features are reshaped to the spatial patch grid.}
    \label{tab:app_eval_features}
    \small
    \renewcommand{\arraystretch}{1.12}
    \begin{tabularx}{\linewidth}{@{}lXX@{}}
        \toprule
        Task & Features & Readout \\
        \midrule
        $k$-NN classification & $\ell_2$-normalized CLS token & Weighted nearest-neighbor vote \\
        Linear / MLP classification & CLS concatenated with mean-pooled patches ($2D$) & Linear / MLP classifier \\
        Semantic segmentation & Patch tokens ($D$ per location) & SyncBatchNorm, dropout, $1\times1$ convolution \\
        Depth estimation & Patch tokens ($D$ per location) & $1\times1$ convolution to 256 depth bins \\
        \bottomrule
    \end{tabularx}
\end{table}

Unless stated otherwise, trained probes use 3 seeds and results are reported as mean $\pm$ standard deviation across these seeds. The deterministic $k$-NN evaluation is reported without error bars. Layer-wise curves and additional evaluations use a single probe seed unless stated otherwise.

\textbf{Data and model selection.}~
Probes are trained on the respective training splits and evaluated on ImageNet-1k validation, ADE20K validation, and the 653-image NYUv2 test set. The same evaluation splits are used for selecting $k$, the classification probe learning rate, and the segmentation checkpoint. Images use ImageNet RGB normalization, expressed on the $[0,1]$ scale for classification and depth and the equivalent $[0,255]$ scale for segmentation.

\subsubsection{Classification}
\label{app:eval_classification}

\textbf{Dataset and metric.}~
ImageNet-1k \citep{deng2009imagenet} contains 1,000 classes, with 1,281,167 training and 50,000 validation images. We use the full training split for both $k$-NN and linear probing and report top-1 accuracy on the validation split.

\textbf{$k$-NN classification.}~
We extract a feature bank from all training images and classify validation images using cosine similarity between normalized CLS features. Both splits use resizing to a short side of 256 followed by a $224\times224$ center crop. Class scores are obtained by summing neighbor-label votes weighted by softmax-normalized cosine similarities at temperature 0.07, following the original DINOv3 $k$-NN evaluation pipeline. We predict the highest-scoring class and report top-1 accuracy. We evaluate $k\in\{10,20,100,200\}$ and report the highest validation accuracy over these choices for each model and evaluated layer.

\textbf{Linear probing.}~
A single-layer linear classifier maps the concatenated CLS and mean-pooled patch features to 1,000 classes, without an additional normalization layer in the head. Training uses random resized cropping to $224\times224$ and horizontal flipping, and validation uses the same center-crop preprocessing as $k$-NN. We optimize cross-entropy with the settings in \cref{tab:app_probe_optimization}. Seven heads are trained in parallel on the same frozen features, one per learning rate, and the best final validation accuracy is reported. The learning rates are

\begin{equation*}
\{2\!\times\!10^{-4},10^{-3},2\!\times\!10^{-3},10^{-2},2\!\times\!10^{-2},10^{-1},2\!\times\!10^{-1}\},
\end{equation*}

Each head is trained for 25,000 steps.

\begin{table}[t]
    \centering
    \caption{Hyperparameters used for the downstream heads. The classification learning-rate grid is specified in \cref{app:eval_classification}.}
    \label{tab:app_probe_optimization}
    \small
    \setlength{\tabcolsep}{4pt}
    \renewcommand{\arraystretch}{1.12}
    \begin{tabularx}{\linewidth}{@{}lXXX@{}}
        \toprule
        Setting & Classification & Segmentation & Depth Estimation \\
        \midrule
        Optimizer & AdamW & AdamW & AdamW \\
        Learning rate & Seven-value grid & $4\times10^{-3}$ & $6\times10^{-4}$ \\
        Weight decay & 0 & $10^{-3}$ & $10^{-4}$ \\
        Adam betas & $(0.9,0.999)$ & $(0.9,0.999)$ & $(0.9,0.99)$ \\
        Batch size & 512 & 64 & 64 \\
        Training steps & 25,000 & 10,000 & 2,400 \\
        LR schedule & Cosine & WarmupOneCycleLR & WarmupOneCycleLR \\
        Warmup steps & None & 375 & 800 \\
        Gradient clipping & None & None & 35.0 \\
        Loss & Cross-entropy & Cross-entropy & SigLoss \\
        Reported checkpoint & Final, best LR & Best validation mIoU & Final \\
        \bottomrule
    \end{tabularx}
\end{table}

\subsubsection{Semantic Segmentation}
\label{app:eval_segmentation}

\textbf{Dataset and metric.}~
We evaluate semantic segmentation on ADE20K \citep{zhou2019ade20k}, which contains indoor and outdoor scenes annotated with 150 semantic classes. We train on its training split and report mean intersection over union (mIoU) on the 2,000-image validation split. Background label zero is remapped to the ignore index, and ignored pixels are excluded from the loss and metrics. Class IoUs are computed from intersection and union counts accumulated over the full validation set and then averaged over the 150 classes.

\textbf{Training and inference.}~
We use the reference linear segmentation head shown in \cref{tab:app_eval_features}, with dropout 0.1 and cross-entropy loss. Training follows DINOv3's ADE20K augmentation pipeline with $512\times512$ crops, scale jitter, horizontal flipping, and photometric distortion. Full hyperparameter settings are listed in \cref{tab:app_probe_optimization}. Validation is performed every 2,500 iterations, and we report the highest validation mIoU obtained during training. Inference resizes images to a short side of 512 and uses sliding windows of size 512 and stride 341, without test-time augmentation.

\subsubsection{Depth Estimation}
\label{app:eval_depth}

\textbf{Dataset and metric.}~
We evaluate monocular depth estimation on NYUv2 \citep{silberman2012nyu}, an indoor RGB-D dataset. We train on the 24,230-frame training list and evaluate on the 653-image test list. We report RMSE in meters, computed separately for each image over its valid pixels and then averaged across images.

\textbf{Training and inference.}~
The downstream model is a linear binned depth head: a $1\times1$ convolution predicts 256 values per pixel. We apply ReLU, add 0.1 to each value, and normalize by their sum, and therefore, predicted depth is the weighted mean of 256 linearly spaced bin centers in $[0.001,10]$ meters. The encoder remains frozen, and the head is trained with the SigLoss implementation. Training uses distributed shuffling and follows DINOv3's preprocessing and augmentation pipeline, including the fixed crop, random $416\times544$ crops, rotation, horizontal flipping, and brightness augmentation. The head uses the hyperparameter settings in \cref{tab:app_probe_optimization}. We evaluate the final checkpoint using the image and its horizontal flip, and average the aligned predictions. Predictions are clamped to $[0.001,10]$ meters.

\subsection{Computational Pattern Analyses}
\label{app:computational_analyses}

\textbf{Data and notation.}~
Let $\widetilde A$ denote a head's post-softmax attention over all tokens. For spatial queries and keys, define the row-normalized patch-to-patch attention

\begin{equation}
     A_{qk}=\frac{\widetilde A_{qk}}{\sum_{j\in\mathcal P}\widetilde A_{qj}},
    \qquad q,k\in\mathcal P,
\end{equation}

where $\mathcal P$ contains the $N$ patch tokens. Spatial routing metrics use $A$, and attention mass metrics use the original $\widetilde A$. Attention statistics use a fixed 12,419-image subset of ImageNet-1k validation set, where images are resized to a short side of 256 and center-cropped to $224\times224$.

\textbf{Attention distance and spatial entropy.}~
For each spatial query $q$, let $p_q$ denote its two-dimensional
patch-grid coordinates.
We compute the expected Euclidean distance to attended spatial
keys, normalized by the maximum distance $d_{\max}$ between
patch-grid positions, and the normalized attention entropy as:

\begin{equation}
    D_q=\sum_{k\in\mathcal P}A_{qk}
    \frac{\|p_k-p_q\|_2}{d_{\max}},
    \qquad
    H_q=-\frac{\sum_{k\in\mathcal P}A_{qk}\log A_{qk}}
    {\log N},
\end{equation}

where $\mathcal P$ is the set of spatial tokens and $N=|\mathcal P|$. Both statistics are averaged over spatial queries and images for each head. Distance measures spatial extent, whereas entropy measures diffuseness: uniform attention has entropy one and one-hot attention has entropy zero. Layer-wise curves average the head-level statistics within each layer.

\textbf{Query--key normalized mutual information.}~
We measure how strongly the distribution of attended keys
depends on the query location. Treating attention as a joint
distribution over query and key locations, we assign equal
probability to each spatial query:
\[
    p(q,k)=\frac{A_{qk}}{N},
    \qquad
    p_K(k)=\frac{1}{N}\sum_q A_{qk}.
\]
Here, $p_K$ is the key distribution averaged over queries.
Using unnormalized Shannon entropy $\mathcal H$, we compute

\begin{equation}
    \operatorname{NMI}(Q;K)=
    \max\left\{0,
    \frac{
        \mathcal H(p_K)
        -\frac{1}{N}\sum_q\mathcal H(A_{q\cdot})
    }{
        \sqrt{\log N\,\mathcal H(p_K)}
    }
    \right\}.
\end{equation}

The numerator measures the reduction in uncertainty about
the attended key when the query location is known.
The denominator normalizes by the geometric mean of the
query and key entropies.

NMI is zero when all queries share the same attention
distribution, whether that distribution is diffuse or
concentrated. Higher values indicate more query-specific
attention, reaching one when each query attends exclusively
to a distinct key. Thus, NMI measures query dependence,
rather than spatial locality or attention concentration.
We compute one value per image and head, then average
over images and heads.

\textbf{CLS--patch attention exchange.}~
We measure the fraction of raw attention allocated from patch queries to the CLS key, and from the CLS query to patch keys:

\begin{equation}
     M_{\mathrm{patch}\to\mathrm{CLS}}
     =\frac{1}{N}\sum_{q\in\mathcal P}\widetilde A_{q,\mathrm{CLS}},
    \qquad
     M_{\mathrm{CLS}\to\mathrm{patch}}
     =\sum_{k\in\mathcal P}\widetilde A_{\mathrm{CLS},k}.
\end{equation}

The masses are averaged over images and then heads. They retain the original attention normalization and therefore quantify the budget assigned to CLS--patch exchange.

\textbf{Layer-skip interventions.}~
We use 512 randomly sampled images from the ImageNet validation set. For a clean block update $h_{s+1}=B_s(h_s)$, skipping block $s$ replaces its entire output with its input, $\bar h^{(s)}_{s+1}=\bar h^{(s)}_s$, removing both the attention and MLP contributions. All other blocks operate normally. Here, $\bar h^{(s)}_j$ denotes the residual stream at position $j$ when block $s$ is skipped. To measure how the resulting representation change propagates, we compute its relative magnitude at subsequent stream positions $j>s$:
\begin{equation}
    \delta_{\mathrm{stream}}(s,j)
    =\mathbb E_{x,t}
    \frac{\|h_{j,t}-\bar h^{(s)}_{j,t}\|_2}
         {\|h_{j,t}\|_2},
\end{equation}

where $x$ indexes images and $t$ indexes tokens. Ratios are computed per token before averaging over tokens and images. We exclude CLS token for this analysis.

\textbf{Qualitative feature maps.}~
We visualize spatial feature similarities for a grown model and its corresponding fixed-depth baseline. We provide an example for the main text and four additional examples for the appendix. At encoder layers 4, 8, 12, 16, 20, and 24, we show CLS--patch similarity between the CLS vector and each patch vector, and patch-to-patch similarity between a selected anchor patch and all patches. Both map types use cosine similarity between features extracted before the backbone's final LayerNorm. We subtract the per-image mean patch vector from all patch features, including the anchor used for patch-to-patch maps. For CLS--patch maps, we compare the uncentered CLS vector with these centered patch features. The displayed anchors are selected deterministically from object interiors using the corresponding ImageNet-S annotations \citep{gao2022imagenets}. We use the same input image and spatial anchor across models and layers.

\subsection{Additional Evaluations}
\label{app:additional_evaluations}

\subsubsection{Robustness to missing evidence}

To test whether growing changes robustness to missing visual evidence and reliance on foreground objects versus their context, we evaluate models under random and semantic input masking on classification and segmentation tasks. Selected pixels are replaced with the ImageNet mean, equivalent to zero after normalization. The backbone and previously trained task heads remain frozen throughout evaluation. Both classification and segmentation follow the same pipeline described in \cref{app:downstream_evaluation}. We use final-layer readouts for the masking comparisons. The results are provided in \cref{app:additional_results_masking}.

Both random and semantic classification masking use the same 12,419 unique ImageNet-S validation images, which provide foreground annotations for the classified object. Although this subset covers 919 classes, we retain the full 1,000-way ImageNet classifier and take the argmax over all logits. Targets use the ImageNet-S class assignments expressed in the ImageNet-1k index space. We therefore report ImageNet-S top-1 accuracy with a 1,000-way head. Segmentation uses the 2,000-image ADE20K validation split.

\textbf{Random masking.}
For random masking, we sample a uniform permutation of the patch indices and mask the first $\operatorname{round}(rN)$ patches, where $N$ is the number of patches and $r\in\{0,0.1,\ldots,1.0\}$ is the masking ratio. Masks are expanded to pixel resolution using nearest-neighbor interpolation. A deterministic per-image permutation is reused across ratios and models with the same patch grid. Thus, increasing the masking ratio progressively removes patches from the same ordering, and each grown model receives the same masks as its fixed-depth baseline. Each setting uses one mask realization per image.

\textbf{Semantic masking.}
Semantic masking either removes the foreground region or retains only that region and masks its context. For classification, foreground masks come from ImageNet-S segmentation annotations, with background and boundary pixels excluded from the foreground. Masks are resized using nearest-neighbor interpolation and center-cropped to match the image shape. For ADE20K, we select the largest class region occupying between 1\% and 90\% of the full native-resolution image, including background and ignored pixels in the area denominator. Selection operates on raw labels 1--150, excluding background label 0 and ignore label 255 before label reduction. Ties favor the smaller class index, and the selected region includes all components of that class. Regions are resized using nearest-neighbor interpolation without snapping them to patch boundaries.

We report top-1 accuracy and mIoU for classification and segmentation tasks, respectively. Segmentation scores are computed over the whole image. Whole-image relative retention is defined as $100P_{\mathrm{masked}}/P_{\mathrm{clean}}$, where the clean reference is the random-masking score at $r=0$, computed with the same model and head. Both classification and segmentation masking use the same evaluated images as their respective clean references. 

\subsubsection{MAE MLP probes}

To test whether the differences observed with linear probing persist when the readout can capture nonlinear relationships, we train MLP classifiers on frozen representations. Two- and three-layer MLP probes replace the linear classifier while retaining its $2D$-dimensional CLS-plus-pooled-patch input. Hidden layers have width $2D$ and GELU activation, without normalization layers. All other settings, including the learning-rate sweep, AdamW optimization, and 25,000-step budget, follow linear probing. The results are shown in \cref{app:mae_adaptation}.

\subsubsection{MAE fine-tuning}

To examine whether the benefits of growing persist when the representations are adapted to the downstream task, we fine-tune MAE variants end-to-end. Following the original MAE fine-tuning recipe, we update the full MAE encoder and attach global mean pooling over patch tokens, a normalization layer, and a linear classifier. We train on ImageNet-1k for 50 epochs with global batch size 1024, using AdamW with weight decay 0.05 and layer-wise learning-rate decay 0.75. The effective peak learning rate is $4\times10^{-3}$, obtained by scaling the base rate $10^{-3}$ by the batch size divided by 256. Five warmup epochs are followed by cosine decay to $10^{-6}$. Training uses random resized crops to $224\times224$, horizontal flips, RandAugment, random erasing with probability 0.25, mixup strength 0.8, CutMix strength 1.0, label smoothing 0.1, and stochastic depth of 0.2 for ViT-L and 0.3 for ViT-H. Evaluation uses resizing to 256 followed by a $224\times224$ center crop and ImageNet normalization, as in linear probing. We report the best validation top-1 accuracy during training. The results are provided in \cref{app:mae_adaptation}.

\subsubsection{Representation similarity}

To examine how growing changes the similarity of representations across depth, we compare layer pairs within each grown model and its fixed-depth baseline using cosine similarity. These comparisons characterize whether layers produce closely aligned features or retain similar relationships among image-token representations. We sample 10,000 ImageNet training images without replacement and apply the classification center-crop preprocessing. At every evaluated depth, the model's single final LayerNorm is applied to the intermediate block output. For each pair of layers (i, j), features are arranged as matrices $X_i,X_j\in\mathbb R^{N\times D}$ with matching image-token pairs in corresponding rows. We compute mean row-wise cosine similarity as the similarity metric. Separate layer-by-layer matrices are computed for the CLS token and patch tokens. See \cref{app:representation_weight_similarity} for the results.

\subsubsection{Weight similarity}

To examine how parameters evolve after growth and whether copied blocks diverge from their sources, we compare MAE checkpoints across training and source--copy pairs within checkpoints. For DINOv3, we assess weight similarity of the same checkpoint. These comparisons use checkpoint parameters without image inputs or trained heads. We use cosine similarity as the similarity metric. Block-level similarities are computed from the concatenation of all parameters in that block rather than by averaging per-tensor metrics. The results are provided in \cref{app:representation_weight_similarity}.

\section{Additional Results}
\label{app:additional_results}

\subsection{Ablation Study}
\label{app:ablation_study}

Here we examine how the benefits of growing depend on the growth strategy, growing schedule, and training configuration. These ablations complement the main results by identifying which choices consistently improve downstream performance and where they introduce task-dependent trade-offs. Training and evaluation protocols are described in \cref{app:training_setup} and \cref{app:evaluation_setup}. We specify the configurations in each section.

\subsubsection{Ablations on Growing Schedule and Strategy}
\label{app:ablation_growing_schedule_strategy}

\begin{table}[t]
    \centering
    \caption{
        Ablations on growing strategy with fixed Prop-1 growing schedule. Each family is compared against its own fixed-depth baseline at matched steps and settings. Linear accuracy, mIoU, and RMSE are reported as mean $\pm$ standard deviation over three downstream seeds, while $k$-NN is deterministic. FLOPs are total pretraining cost relative to the matched baseline. \textbf{Bold} and \underline{underline} mark the best and second best within each family.
    }
    \label{tab:app_ablation_strategies}
    \small
    \setlength{\tabcolsep}{4.2pt}
    \renewcommand{\arraystretch}{1.15}
    \begin{tabular}{lccccc}
        \toprule
        & Training & \multicolumn{2}{c}{ImageNet-1k}
        & ADE20K & NYU \\
        \cmidrule(lr){3-4}
        Model & FLOPs (\%) $\downarrow$
        & $k$-NN $\uparrow$ & Linear $\uparrow$
        & mIoU $\uparrow$ & RMSE $\downarrow$ \\
        \midrule
        \multicolumn{6}{l}{\textit{MAE ViT-L/16, Prop-1}} \\
        \quad Baseline
        & $100.0$
        & $31.40$
        & $60.28 \pm 0.07$
        & $26.29 \pm 0.03$
        & $0.6393 \pm 0.0009$ \\
        \quad Stacking
        & $79.6$
        & $\underline{33.86}$
        & $\mathbf{63.64 \pm 0.08}$
        & $\mathbf{27.73 \pm 0.04}$
        & $\mathbf{0.5855 \pm 0.0001}$ \\
        \quad MIDAS
        & $79.6$
        & $32.21$
        & $59.59 \pm 0.07$
        & $25.82 \pm 0.08$
        & $\underline{0.6219 \pm 0.0004}$ \\
        \quad LIDAS
        & $79.6$
        & $\mathbf{34.09}$
        & $\underline{61.08 \pm 0.08}$
        & $\underline{26.54 \pm 0.08}$
        & $0.6292 \pm 0.0009$ \\
        \midrule
        \addlinespace[2pt]
        \multicolumn{6}{l}{\textit{MAE ViT-H/14, Prop-1}} \\
        \quad Baseline
        & $100.0$
        & $\underline{36.59}$
        & $\underline{66.25 \pm 0.10}$
        & $28.76 \pm 0.03$
        & $0.5886 \pm 0.0006$ \\
        \quad Stacking
        & $75.2$
        & $\mathbf{37.99}$
        & $66.05 \pm 0.05$
        & $\underline{29.66 \pm 0.03}$
        & $\mathbf{0.5680 \pm 0.0002}$ \\
        \quad MIDAS
        & $75.2$
        & $33.83$
        & $65.82 \pm 0.04$
        & $29.17 \pm 0.03$
        & $0.5761 \pm 0.0002$ \\
        \quad LIDAS
        & $75.2$
        & $34.60$
        & $\mathbf{68.05 \pm 0.11}$
        & $\mathbf{30.18 \pm 0.02}$
        & $\underline{0.5691 \pm 0.0009}$ \\
        \midrule
        \addlinespace[2pt]
        \multicolumn{6}{l}{\textit{DINOv3 ViT-L/16, Prop-1}} \\
        \quad Baseline
        & $100.0$
        & $\mathbf{82.17}$
        & $\mathbf{83.40 \pm 0.04}$
        & $\mathbf{44.70 \pm 0.10}$
        & $\underline{0.4650 \pm 0.0001}$ \\
        \quad Stacking
        & $72.5$
        & $81.24$
        & $82.82 \pm 0.02$
        & $43.45 \pm 0.07$
        & $0.4721 \pm 0.0006$ \\
        \quad MIDAS
        & $72.5$
        & $81.81$
        & $83.15 \pm 0.03$
        & $44.05 \pm 0.08$
        & $\mathbf{0.4625 \pm 0.0005}$ \\
        \quad LIDAS
        & $72.5$
        & $\underline{81.92}$
        & $\underline{83.31 \pm 0.03}$
        & $\underline{44.12 \pm 0.08}$
        & $0.4697 \pm 0.0007$ \\
        \bottomrule
    \end{tabular}
\end{table}

\begin{table}[t]
    \centering
    \caption{
        Growth-schedule ablations. The results are shown as mean $\pm$ standard deviation over three downstream seeds, while $k$-NN is deterministic. \textbf{Bold} and \underline{underline} mark the best and second best within each family.
        }
    \label{tab:app_schedules}
    \small
    \setlength{\tabcolsep}{4.2pt}
    \renewcommand{\arraystretch}{1.15}
    \begin{tabular}{lccccc}
        \toprule
        & Training & \multicolumn{2}{c}{ImageNet-1k} & ADE20K & NYU \\
        \cmidrule(lr){3-4}
        Model & FLOPs (\%) $\downarrow$
        & $k$-NN $\uparrow$ & Linear $\uparrow$
        & mIoU $\uparrow$ & RMSE $\downarrow$ \\
        \midrule
        \multicolumn{6}{l}{\textit{MAE ViT-L, Stacking}} \\
        \quad Baseline & $100.0$ & $31.40$ & $60.28 \pm 0.07$ & $26.29 \pm 0.03$ & $0.6393 \pm 0.0009$ \\
        \quad Prop-1 & $79.6$ & $33.86$ & \underline{$63.64 \pm 0.08$} & $27.73 \pm 0.04$ & $\mathbf{0.5855 \pm 0.0001}$ \\
        \quad Prop-2 & $85.9$ & $\mathbf{36.33}$ & $\mathbf{64.23 \pm 0.05}$ & $\mathbf{28.27 \pm 0.03}$ & $0.6137 \pm 0.0002$ \\
        \quad Prop-3 & $89.7$ & $\underline{35.83}$ & $\mathbf{64.23 \pm 0.05}$ & $\underline{28.04 \pm 0.03}$ & $\underline{0.5963 \pm 0.0002}$ \\
        \midrule
        \addlinespace[2pt]
        \multicolumn{6}{l}{\textit{DINOv3 ViT-L}} \\
        \quad Baseline & $100.0$ & $\underline{82.17}$ & $83.40 \pm 0.04$ & $44.70 \pm 0.10$ & $0.4650 \pm 0.0001$ \\
        \quad MIDAS, Prop-1 & $72.5$ & $81.81$ & $83.15 \pm 0.03$ & $44.05 \pm 0.08$ & $0.4625 \pm 0.0005$ \\
        \quad MIDAS, Prop-2 & $81.0$ & $82.13$ & $\underline{83.72 \pm 0.03}$ & $44.75 \pm 0.04$ & $\mathbf{0.4519 \pm 0.0002}$ \\
        \quad MIDAS, Prop-3 & $86.1$ & $\mathbf{82.23}$ & $83.71 \pm 0.06$ & $\underline{45.64 \pm 0.10}$ & $0.4614 \pm 0.0006$ \\
        \quad LIDAS, Prop-1 & $72.5$ & $81.92$ & $83.31 \pm 0.03$ & $44.12 \pm 0.08$ & $0.4697 \pm 0.0007$ \\
        \quad LIDAS, Prop-2 & $81.0$ & $81.96$ & $83.67 \pm 0.06$ & $45.52 \pm 0.06$ & $0.4653 \pm 0.0002$ \\
        \quad LIDAS, Prop-3 & $86.1$ & $82.16$ & $\mathbf{83.73 \pm 0.04}$ & $\mathbf{45.77 \pm 0.09}$ & $\underline{0.4523 \pm 0.0005}$ \\
        \bottomrule
    \end{tabular}
\end{table}

To distinguish the effect of where blocks are inserted from how long each depth is trained, we first compare growth strategies under the same Prop-1 schedule, then vary the schedule for selected strategies. Within each model family, the strategy comparison uses identical stage durations and therefore identical pretraining FLOPs. The schedule comparison keeps the total number of training steps fixed but changes their allocation across depths: larger exponents $\alpha$ in a Prop-$\alpha$ schedule allocate more steps to deeper stages and consequently increase computation. ViT-L comparisons use cosine learning-rate schedules, while ViT-H comparisons use a WSD schedule. MAE models are trained for 50k steps and DINOv3 ViT-L is trained for 125k steps. The remaining training settings follow \cref{app:training_setup}.

\textbf{Growing strategy.}~
\cref{tab:app_ablation_strategies} compares growing strategies under a common Prop-1 schedule. It repeats the MAE results from \cref{tab:main_downstream} and adds DINOv3 strategy comparisons under Prop-1, complementing the Prop-3 results in the main text. We observe that the best growing strategy depends on model family and scale. For MAE ViT-L, Stacking gives the strongest linear classification, segmentation, and depth results, improving over the baseline by 3.36 percentage points in linear accuracy and 1.44 points in mIoU, while reducing RMSE from 0.6393 to 0.5855. LIDAS achieves the highest $k$-NN accuracy and improves all four metrics over the baseline. MIDAS improves $k$-NN and depth estimation but falls below the baseline on linear classification and segmentation. All three strategies use 79.6\% of the baseline FLOPs, so these differences arise under the same computation budget. The pattern holds at the larger ViT-H scale despite an even larger 24.8\% FLOP reduction. For MAE ViT-H, LIDAS gives the strongest linear classification and segmentation results, improving linear accuracy from 66.25 to 68.05 and mIoU from 28.76 to 30.18. It also reduces RMSE from 0.5886 to 0.5691, although Stacking achieves the best depth and $k$-NN results.

\begin{wrapfigure}{r}{0.42\textwidth}
    \vspace{-0.8em}
    \centering
    \includegraphics[width=\linewidth]{
        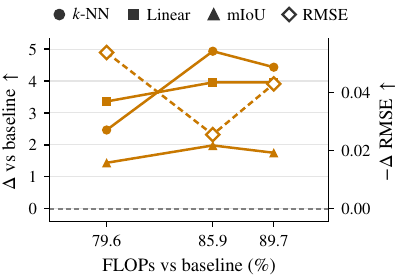
    }
    \captionsetup{font=footnotesize,skip=4pt}
    \caption{
        \textbf{MAE compute--performance trade-off.}
        Downstream gains over the fixed-depth baseline versus
        relative pretraining FLOPs for Stacking with Prop-1,
        Prop-2, and Prop-3 (left to right).
        Accuracy and mIoU gains use the left axis (percentage
        points), and RMSE reduction uses the right axis (meters).
        All three schedules improve performance with less compute,
        while the best schedule depends on the task.
    }
    \label{fig:app_flops_tradeoff_mae}
\end{wrapfigure}

For DINOv3, Prop-1 produces a different pattern. All three strategies use 27.5\% less compute but fall below the baseline on classification and segmentation, although LIDAS comes closest, reaching 83.31\% linear accuracy against 83.40\% for the baseline. MIDAS achieves the lowest depth RMSE, 0.4625 compared with 0.4650 for the baseline, while Stacking performs worst across all four metrics. Based on this comparison, we focus the main DINOv3 analyses on MIDAS and LIDAS and exclude Stacking from the rest of the evaluations.

\textbf{Growing schedule.}~
\cref{tab:app_schedules,fig:app_flops_tradeoff_mae} show that allocating more training to deeper stages can improve performance, but the gains depend on the task. For MAE Stacking, moving from Prop-1 to Prop-2 increases $k$-NN accuracy from 33.86\% to 36.33\%, linear accuracy from 63.64\% to 64.23\%, and mIoU from 27.73\% to 28.27\%. Depth estimation instead favors Prop-1: RMSE increases from 0.5855 to 0.6137 under Prop-2. Prop-3 matches Prop-2's reported linear accuracy and partially recovers depth performance, but has lower $k$-NN accuracy and mIoU despite using more computation. All three schedules outperform the fixed-depth MAE baseline on every task. These gains show that adjusting the growing schedule can further improve classification and segmentation performance beyond the Prop-1 results shown in \cref{tab:main_downstream} at the cost of additional pretraining compute.

For DINOv3, both MIDAS and LIDAS improve classification and segmentation when moving from Prop-1 to Prop-2 or Prop-3. Under Prop-3, MIDAS exceeds the baseline on all four metrics, while LIDAS improves linear accuracy, segmentation, and depth, with $k$-NN accuracy only 0.01 percentage points below the baseline. LIDAS with Prop-3 achieves the highest linear accuracy and mIoU, whereas MIDAS with Prop-2 achieves the lowest RMSE. These configurations use 86.1\% and 81.0\% of baseline FLOPs, respectively.

Together, these results show that growth strategy and schedule both matter, and that no single configuration is best across objectives and tasks. Although allocating more steps to deeper stages increases pretraining FLOPs, it substantially improves the DINOv3 comparisons, while MAE exhibits a clearer trade-off between classification or segmentation and depth estimation.

\subsubsection{Ablations on Training Configuration}
\label{app:ablation_training_configuration}

\begin{table}[t]
    \centering
    \caption{MAE ViT-L learning-rate and schedule ablations. The first block contains fixed-depth sweep runs, and the second block compares Stacking (Prop-1) with the selected cosine ($2.4\times10^{-3}$) and WSD ($10^{-3}$) references. Learning rates are effective optimizer peaks. Results are shown as mean $\pm$ standard deviation over three downstream seeds, while $k$-NN is deterministic. \textbf{Bold} marks the best performance in each block.}
    \label{tab:app_mae_training_ablation}
    \small
    \setlength{\tabcolsep}{3pt}
    \renewcommand{\arraystretch}{1.13}
    \newcolumntype{C}{>{\centering\arraybackslash}c}
    \begin{tabularx}{\linewidth}{@{}XCCCCC@{}}
        \toprule
        & Training & \multicolumn{2}{c}{ImageNet-1k} & ADE20K & NYUv2 \\
        \cmidrule(lr){3-4}
        Configuration & FLOPs (\%) $\downarrow$ & $k$-NN $\uparrow$ & Linear $\uparrow$ & mIoU $\uparrow$ & RMSE $\downarrow$ \\
        \midrule
        \multicolumn{6}{l}{\textit{Fixed depth, 50k steps}} \\
        \quad cosine, 5e-4 & $100.0$ & $31.82$ & $59.62 \pm 0.06$ & $24.22 \pm 0.04$ & $0.6557 \pm 0.0004$ \\
        \quad cosine, 1e-3 & $100.0$ & $34.77$ & $61.60 \pm 0.04$ & $26.02 \pm 0.03$ & $0.6226 \pm 0.0006$ \\
        \quad cosine, 2.4e-3 & $100.0$ & $31.40$ & $60.28 \pm 0.07$ & $26.29 \pm 0.03$ & $0.6393 \pm 0.0009$ \\
        \quad WSD, 5e-4 & $100.0$ & $34.71$ & $61.22 \pm 0.06$ & $25.80 \pm 0.04$ & $0.6381 \pm 0.0002$ \\
        \quad WSD, 1e-3 & $100.0$ & $\mathbf{35.71}$ & $\mathbf{64.02 \pm 0.07}$ & $\mathbf{27.07 \pm 0.03}$ & $\mathbf{0.6214 \pm 0.0003}$ \\
        \quad WSD, 2.4e-3 & $100.0$ & $31.55$ & $60.80 \pm 0.06$ & $25.97 \pm 0.05$ & $0.6316 \pm 0.0004$ \\
        \addlinespace[2pt]
        \multicolumn{6}{l}{\textit{Selected baseline--grown pairs, 50k steps}} \\
        \quad Baseline, cosine, 2.4e-3 & $100.0$ & $31.40$ & $60.28 \pm 0.07$ & $26.29 \pm 0.03$ & $0.6393 \pm 0.0009$ \\
        \quad Stacking, cosine, 2.4e-3 & $79.6$ & $33.86$ & $63.64 \pm 0.08$ & $27.73 \pm 0.04$ & $0.5855 \pm 0.0001$ \\
        \quad Baseline, WSD, 1e-3 & $100.0$ & $35.71$ & $64.02 \pm 0.07$ & $27.07 \pm 0.03$ & $0.6214 \pm 0.0003$ \\
        \quad Stacking, WSD, 1e-3 & $79.6$ & $\mathbf{40.49}$ & $\mathbf{64.54 \pm 0.06}$ & $\mathbf{28.20 \pm 0.04}$ & $\mathbf{0.5803 \pm 0.0002}$ \\
        \bottomrule
    \end{tabularx}
\end{table}

\begin{figure}[t]
    \centering
    \includegraphics[width=\linewidth]{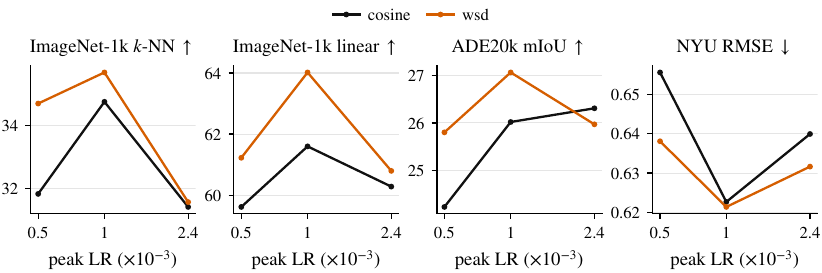}
    \caption{Learning-rate sensitivity of fixed-depth MAE ViT-L trained for 50k steps. Panels show final-layer $k$-NN and linear classification accuracy, ADE20K mIoU, and NYUv2 RMSE under cosine and WSD schedules. Learning rates are effective optimizer peaks.}
    \label{fig:app_lr_mae}
\end{figure}

To examine how the downstream performance depends on the pretraining recipe, we vary learning rates, learning-rate schedules, and selected DINOv3 hyperparameters. The learning-rate sweeps use fixed-depth ViT-L models, and separate comparisons evaluate selected recipes and interventions in grown models. All results use final-layer readouts and the evaluation protocols in \cref{app:evaluation_setup}. DINOv3 ablations were conducted sequentially, with a separate reference configuration for each comparison, rather than as a joint hyperparameter grid.

\textbf{MAE learning rate and schedule.}~
We first compare cosine decay and warmup--stable--decay (WSD) at peak learning rates of $5\times10^{-4}$, $10^{-3}$, and $2.4\times10^{-3}$, training fixed-depth MAE ViT-L for 50k steps. WSD uses a linear cooldown over the final 20\% of training. The main-text baseline retains the original MAE cosine schedule and default base learning rate, giving an effective peak of $2.4\times10^{-3}$. The alternative learning rates and WSD schedules are evaluated here to assess sensitivity to the pretraining recipe. All configurations in this comparison use 1,250 warmup steps. The fixed-depth comparisons and adopted baseline--grown pairs are reported in \cref{tab:app_mae_training_ablation}, with the learning-rate trends shown in \cref{fig:app_lr_mae}.

The strongest fixed-depth configuration across all four metrics is WSD at $10^{-3}$, which achieves 35.71\% $k$-NN accuracy, 64.02\% linear accuracy, 27.07\% mIoU, and 0.6214 RMSE. Increasing the learning rate to $2.4\times10^{-3}$ worsens all four metrics under WSD. With cosine decay, $10^{-3}$ gives the highest $k$-NN and linear accuracy and the lowest RMSE, whereas the adopted $2.4\times10^{-3}$ recipe gives the highest mIoU (26.29\% vs. 26.02\%). Thus, the preferred cosine configuration depends on the downstream task, while WSD at $10^{-3}$ outperforms all listed cosine configurations.

Stacking improves all four metrics over the corresponding baseline under both adopted recipes. Under WSD at $10^{-3}$, the gains are 4.78 percentage points in $k$-NN accuracy, 0.52 percentage points in linear accuracy, and 1.13 points in mIoU, while RMSE decreases from 0.6214 to 0.5803. Both grown configurations use 79.6\% of the corresponding fixed-depth FLOPs. The improvement from growing therefore persists under the stronger WSD reference with matched training settings.

\begin{figure}[t]
    \centering
    \includegraphics[width=\linewidth]{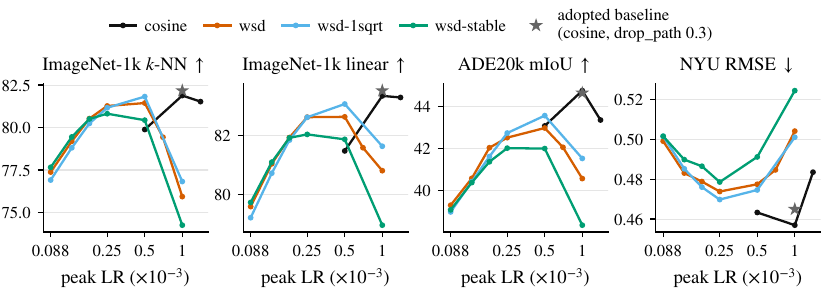}
    \caption{Learning-rate and schedule sensitivity of fixed-depth DINOv3 ViT-L trained for 125k steps. The horizontal axis shows peak learning rates before batch-size scaling. In the legend, \texttt{wsd} uses linear cooldown, \texttt{wsd-1sqrt} uses $1-\sqrt{s}$ cooldown, and \texttt{wsd-stable} has no cooldown. Stars identify the adopted cosine baseline with stochastic depth 0.3. Results are final-layer evaluations without uncertainty bands.}
    \label{fig:app_lr_dinov3}
\end{figure}

\begin{table}[t]
    \centering
    \caption{
    DINOv3 ViT-L configuration ablations.
    Entries are variant minus the local reference.
    Fixed-depth stochastic-depth comparisons use a peak learning rate of
    $5\times10^{-4}$ for WSD variants
    and $10^{-3}$ for cosine.
    The fixed-depth schedule comparison uses linear WSD as its
    reference, with a peak learning rate $5\times10^{-4}$
    and stochastic depth 0.1.
    All grown models use Prop-1.
    Within each strategy, the grown stochastic-depth and
    layer-wise learning-rate ablations share one reference run:
    WSD with $5\times10^{-4}$,
    $1-\sqrt{s}$ cooldown,
    stochastic depth 0.1, and layer-wise learning-rate decay 0.9.
    This reference recipe is maintained across all growth stages.
    These variants change only the indicated hyperparameter
    in the final growth stage.
    The separate grown cooldown comparison replaces
    $1-\sqrt{s}$ with cosine cooldown within WSD,
    holding the peak learning rate $5\times10^{-4}$
    and stochastic depth 0.1 fixed. \textbf{Bold} marks improvement, which is negative for RMSE.
    }
    \label{tab:app_dinov3_knobs}
    \small
    \setlength{\tabcolsep}{3pt}
    \renewcommand{\arraystretch}{1.13}
    \newcolumntype{C}{>{\centering\arraybackslash}c}
    \begin{tabularx}{\linewidth}{@{}XCCCC@{}}
        \toprule
        & \multicolumn{2}{c}{ImageNet-1k} & ADE20K & NYUv2 \\
        \cmidrule(lr){2-3}
        Configuration & $k$-NN $\uparrow$ & Linear $\uparrow$ & mIoU $\uparrow$ & RMSE $\downarrow$ \\
        \midrule
        \multicolumn{5}{l}{\textit{Fixed depth: stochastic depth $0.1 \rightarrow 0.3$}} \\
        \quad WSD, linear & $-0.77$ & $-0.57$ & $-0.71$ & $+0.0036$ \\
        \quad WSD, $1-\sqrt{s}$ & $-0.53$ & $-0.50$ & $-0.86$ & $\mathbf{-0.0010}$ \\
        \quad No cooldown & $-1.12$ & $-0.88$ & $-1.08$ & $+0.0051$ \\
        \quad Cosine & $\mathbf{+0.23}$ & $\mathbf{+0.18}$ & $-0.18$ & $+0.0003$ \\
        \addlinespace[2pt]
        \multicolumn{5}{l}{\textit{Fixed depth: schedule (reference: linear WSD)}} \\
        \quad WSD, $1-\sqrt{s}$ & $\mathbf{+0.38}$ & $\mathbf{+0.43}$ & $\mathbf{+0.59}$ & $\mathbf{-0.0028}$ \\
        \quad WSD, no cooldown & $-1.01$ & $-0.75$ & $-0.98$ & $+0.0137$ \\
        \quad Cosine (full schedule) & $-1.56$ & $-1.14$ & $\mathbf{+0.10}$ & $\mathbf{-0.0141}$ \\
        \addlinespace[2pt]
        \multicolumn{5}{l}{\textit{Grown: stochastic depth $0.1 \rightarrow 0.3$, final stage}} \\
        \quad MIDAS & $-0.32$ & $-0.38$ & $-0.20$ & $\mathbf{-0.0114}$ \\
        \quad LIDAS & $-0.35$ & $-0.50$ & $\mathbf{+0.07}$ & $+0.0035$ \\
        \addlinespace[2pt]
        \multicolumn{5}{l}{\textit{Grown: layer-wise LR decay $0.9 \rightarrow 0.95$, final stage}} \\
        \quad MIDAS & $-0.38$ & $-0.22$ & $-0.22$ & $\mathbf{-0.0097}$ \\
        \quad LIDAS & $-0.37$ & $-0.56$ & $\mathbf{+0.01}$ & $+0.0010$ \\
        \addlinespace[2pt]
        \multicolumn{5}{l}{\textit{Grown: layer-wise LR decay $0.9 \rightarrow 0.99$, final stage}} \\
        \quad MIDAS & $-1.49$ & $-1.51$ & $-1.22$ & $\mathbf{-0.0146}$ \\
        \quad LIDAS & $-1.66$ & $-1.77$ & $-0.74$ & $+0.0142$ \\
        \addlinespace[2pt]
        \multicolumn{5}{l}{\textit{Grown: WSD cooldown $1-\sqrt{s} \rightarrow$ cosine}} \\
        \quad MIDAS & $-0.34$ & $-0.15$ & $-0.11$ & $\mathbf{-0.0040}$ \\
        \quad LIDAS & $-0.38$ & $-0.29$ & $-0.02$ & $+0.0004$ \\

        \bottomrule
    \end{tabularx}
\end{table}

\textbf{DINOv3 learning rate and schedule.}~
For fixed-depth DINOv3 trained for 125k steps, we compare cosine decay with WSD using linear or $1-\sqrt{s}$ cooldowns, as well as a warmup--stable schedule without cooldown, where $s$ denotes normalized cooldown progress. In \cref{fig:app_lr_dinov3}, the plotted learning rates are peak learning rates before batch-size scaling. At global batch size 2048, the effective optimizer peak is $4\sqrt{2}\,lr$. We keep all other hyperparameters fixed as described in \cref{app:training_setup}.

The favorable learning-rate range differs across schedules. Cosine decay reaches its strongest classification and segmentation results near $10^{-3}$, whereas the WSD curves favor lower values, around $3.5\times10^{-4}$ to $5\times10^{-4}$ depending on the task. Raising the WSD learning rate to $10^{-3}$ substantially reduces classification and segmentation performance and increases depth error, with the largest deterioration occurring without cooldown. 
Replacing linear cooldown with $1-\sqrt{s}$ improves all four metrics. Removing the cooldown worsens all four, including reductions of 0.75 points in linear accuracy and 0.98 points in mIoU and an increase of 0.0137 in RMSE. Switching to a full cosine schedule reduces linear accuracy by 1.14 points but improves mIoU by 0.10 points and lowers RMSE by 0.0141. These comparisons show that cooldown improves the tested WSD recipes, while the preferred schedule and learning rate depend on the downstream task.

\textbf{DINOv3 stochastic depth and layer-wise learning rates.}~
The paired changes in \cref{tab:app_dinov3_knobs} show that stronger stochastic depth is not uniformly beneficial. Increasing its rate from 0.1 to 0.3 reduces classification and segmentation performance under all three listed WSD variants. With cosine decay, it instead improves $k$-NN and linear accuracy by 0.23 and 0.18 points, while slightly worsening segmentation and depth. This supports treating stochastic depth as part of the training recipe rather than transferring its preferred value across schedules automatically.

For grown MIDAS and LIDAS models, the stochastic-depth and layer-wise learning-rate ablations use the same WSD reference within each strategy, as specified in \cref{tab:app_dinov3_knobs}. Each variant changes only the indicated hyperparameter in the final growth stage, allowing its effect to be assessed relative to the unchanged reference recipe. Increasing stochastic depth to 0.3 reduces both classification scores for both strategies. MIDAS improves depth RMSE by 0.0114, whereas LIDAS has slightly worse depth and slightly higher mIoU. Increasing the layer-wise learning-rate decay factor from 0.9 toward 1 reduces the attenuation of learning rates in earlier layers. At 0.99, linear accuracy falls by 1.51 for MIDAS and 1.77 percentage points for LIDAS, and mIoU falls by 1.22 and 0.74 points. Depth estimation again responds differently: MIDAS improves by 0.0146 RMSE, while LIDAS worsens by 0.0142. The grown cooldown ablation compares $1-\sqrt{s}$ with cosine cooldown within WSD, holding stochastic depth at 0.1 on both sides. Cosine cooldown reduces both classification scores and mIoU for MIDAS and LIDAS. It lowers MIDAS depth RMSE by 0.0040 but slightly increases LIDAS RMSE by 0.0004. Thus, this change provides no improvement shared across tasks, reinforcing the strategy-dependent responses observed in the other configuration ablations.

\begin{figure}[t]
    \centering
    \includegraphics[width=\linewidth]{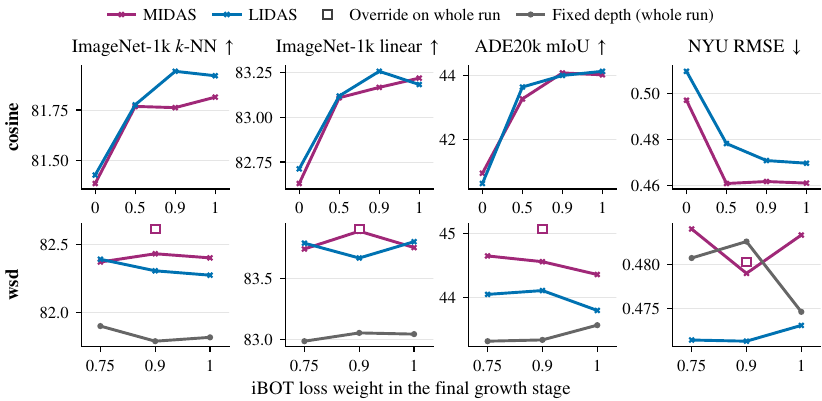}
    \caption{
    Sensitivity to the iBOT loss weight for DINOv3 ViT-L.
    Columns show final-layer $k$-NN accuracy, linear accuracy,
    ADE20K mIoU, and NYUv2 RMSE, and rows show cosine and WSD schedules.
    MIDAS and LIDAS use Prop-1, with the indicated weight applied
    only in the final growth stage, and weight 1.0 in earlier stages.
    Grey fixed-depth curves in the WSD row apply each weight
    throughout training. Hollow squares denote MIDAS with
    weight 0.9 throughout all stages as a reference point.
    }
    \label{fig:app_ibot_dinov3}
\end{figure}

\textbf{DINOv3 iBOT loss weight.}~
To examine sensitivity to patch-level supervision after the final expansion, we vary the iBOT loss weight only in the final growth stage of DINOv3, with results in \cref{fig:app_ibot_dinov3}. The weight remains 1.0 throughout the earlier stages. Consequently, a weight of zero disables the iBOT loss only during the final stage.

For cosine-trained grown models, disabling this loss in the final stage gives the weakest performance on all four metrics. Increasing the weight toward 1 raises mIoU from approximately 41 to 44 for both MIDAS and LIDAS. MIDAS depth RMSE falls from approximately 0.497 at weight zero to 0.461 at weight 0.5, with little further change. LIDAS improves more gradually, from approximately 0.510 to 0.470 across the displayed range. Classification also improves from zero to 0.5, followed by smaller changes near weights 0.9 and 1. These results indicate that continued iBOT supervision remains beneficial after the final growing stage.

Under WSD, varying the final-stage weight between 0.75 and 1 produces smaller, non-monotonic changes. MIDAS achieves its highest classification scores at 0.9 but its highest mIoU at 0.75, while LIDAS favors different weights across tasks. The fixed-depth curves apply each weight throughout training, so their intervention timing differs from that of the grown curves.

The additional MIDAS control at weight 0.9 directly compares applying the override throughout training with applying it only in the final stage. Using 0.9 throughout training gives 82.6\% versus 82.4\% $k$-NN accuracy and 45.1\% versus 44.6\% mIoU. Linear accuracy changes little, while RMSE is slightly higher, 0.480 versus 0.479. Thus, applying the same weight earlier in training changes the observed trade-off: the whole-run override improves $k$-NN accuracy and segmentation in this comparison, but does not improve every task.

\begin{table}[t]
    \centering
    \caption{MAE ViT-L after 500k training steps. Stacking uses Prop-1 with growth horizon $T_g$ set to 50k or 500k steps. The learning-rate schedule spans the full 500k-step budget in both cases. FLOPs are relative to the 500k-step fixed-depth baseline. Results are shown as mean $\pm$ standard deviation over three downstream seeds, while $k$-NN is deterministic. \textbf{Bold} marks the best performance within each group.}
    \label{tab:app_mae_growth_horizon}
    \small
    \setlength{\tabcolsep}{3pt}
    \renewcommand{\arraystretch}{1.13}
    \newcolumntype{C}{>{\centering\arraybackslash}c}
    \begin{tabularx}{\linewidth}{@{}XCCCCC@{}}
        \toprule
        & Training & \multicolumn{2}{c}{ImageNet-1k} & ADE20K & NYUv2 \\
        \cmidrule(lr){3-4}
        Configuration & FLOPs (\%) $\downarrow$ & $k$-NN $\uparrow$ & Linear $\uparrow$ & mIoU $\uparrow$ & RMSE $\downarrow$ \\
        \midrule
        \multicolumn{6}{l}{\textit{Cosine, peak LR $2.4\times10^{-3}$}} \\
        \quad Baseline & $100.0$ & $49.96$ & $75.24 \pm 0.04$ & $32.34 \pm 0.04$ & $0.5074 \pm 0.0008$ \\
        \quad Stacking, $T_g=50$k & $98.0$ & $51.86$ & $\mathbf{76.40 \pm 0.07}$ & $\mathbf{33.27 \pm 0.07}$ & $\mathbf{0.4981 \pm 0.0001}$ \\
        \quad Stacking, $T_g=500$k & $79.6$ & $\mathbf{57.97}$ & $75.14 \pm 0.06$ & $32.93 \pm 0.03$ & $0.5040 \pm 0.0007$ \\
        \addlinespace[2pt]
        \multicolumn{6}{l}{\textit{WSD, peak LR $10^{-3}$}} \\
        \quad Baseline & $100.0$ & $47.01$ & $74.60 \pm 0.05$ & $31.64 \pm 0.06$ & $0.5234 \pm 0.0003$ \\
        \quad Stacking, $T_g=50$k & $98.0$ & $48.83$ & $74.81 \pm 0.05$ & $32.84 \pm 0.04$ & $\mathbf{0.5030 \pm 0.0004}$ \\
        \quad Stacking, $T_g=500$k & $79.6$ & $\mathbf{55.22}$ & $\mathbf{75.12 \pm 0.07}$ & $\mathbf{33.43 \pm 0.03}$ & $0.5090 \pm 0.0004$ \\
        \bottomrule
    \end{tabularx}
\end{table}

\subsubsection{Training Budget and Growth Horizon}
\label{app:ablation_training_growth_horizon}

To examine how additional training and earlier attainment of full depth affect downstream performance, we vary the total training budget and the horizon over which the growth schedule is defined. These comparisons extend the main MAE training study in \cref{fig:mae_extended_training}. The total training horizon determines the learning-rate schedule, while the growth horizon determines how training steps are allocated across depths.

\textbf{Growing MAE over 50k or 500k steps.}~
We train MAE ViT-L for 500k steps and compare Stacking with Prop-1 defined over either the first 50k steps or the full 500k-step budget. The former reaches full depth early and then continues training, as in the main text, and the latter spends a larger fraction of training at reduced depth. Their total pretraining costs are 98.0\% and 79.6\% of the 500k-step fixed-depth baseline, respectively. The encoder's final architecture and the fixed decoder are shared.

With cosine learning rate schedule, early growth improves all four metrics over the baseline, reaching 76.40\% linear accuracy and 0.4981 RMSE in \cref{tab:app_mae_growth_horizon}. Spreading growth over 500k steps yields much higher $k$-NN accuracy, 57.97\% compared with 51.86\% for early growth and 49.96\% for the baseline. However, its linear accuracy is 75.14\%, slightly below the baseline's 75.24\%, and its segmentation and depth scores are weaker than those of early growth. Under WSD, growing over 500k steps improves all four metrics over the fixed-depth baseline. Compared with early growth, it improves $k$-NN, linear accuracy, and mIoU, but has higher RMSE (0.5090 versus 0.5030). Thus, distributing growth across the full training budget retains substantially more compute savings and strongly favors $k$-NN performance in both recipes, while downstream gains on other tasks remain schedule-specific.

\begin{table}[t]
    \centering
    \caption{DINOv3 training-budget and growth-horizon ablations. Metric entries are changes relative to the corresponding 125k-step Prop-1 grown reference, whose growth horizon is 125k. Training extension preserves the original growth boundaries, whereas the 100k-horizon experiment retains 125k training steps and changes those boundaries. \textbf{Bold} marks improvement.}
    \label{tab:app_dinov3_horizon}
    \small
    \setlength{\tabcolsep}{3pt}
    \renewcommand{\arraystretch}{1.13}
    \newcolumntype{C}{>{\centering\arraybackslash}c}
    \begin{tabularx}{\linewidth}{@{}XCCCCC@{}}
        \toprule
        & Training & \multicolumn{2}{c}{ImageNet-1k} & ADE20K & NYUv2 \\
        \cmidrule(lr){3-4}
        Configuration & FLOPs (\%) $\downarrow$ & $k$-NN $\uparrow$ & Linear $\uparrow$ & mIoU $\uparrow$ & RMSE $\downarrow$ \\
        \midrule
        \multicolumn{6}{l}{\textit{WSD: total training budget $125$k $\rightarrow 160$k}} \\
        \quad MIDAS & $100$ & $-0.63$ & $-0.03$ & $\mathbf{+0.13}$ & $+0.0070$ \\
        \quad LIDAS & $100$ & $-0.87$ & $-0.15$ & $\mathbf{+0.13}$ & $+0.0064$ \\
        \addlinespace[2pt]
        \multicolumn{6}{l}{\textit{MIDAS: growth horizon $125$k $\rightarrow 100$k}} \\
        \quad Cosine & $78.0$ & $\mathbf{+0.29}$ & $\mathbf{+0.27}$ & $\mathbf{+0.23}$ & $+0.0049$ \\
        \quad WSD & $78.0$ & $\mathbf{+0.12}$ & $\mathbf{+0.21}$ & $\mathbf{+0.53}$ & $\mathbf{-0.0001}$ \\
        \bottomrule
    \end{tabularx}
\end{table}

\textbf{DINOv3 training extension versus earlier growth.}~
For DINOv3, we first extend WSD training from 125k to 160k steps to match the fixed-depth baseline FLOPs, while preserving the original Prop-1 growth boundaries. Only the final full-depth stage is extended, and the 25k-step cooldown is shifted to the end of the longer run. This uses the same amount of FLOPs as the 125k-step fixed-depth baseline. Relative to the corresponding 125k-step grown models, both MIDAS and LIDAS gain 0.13 mIoU, but lose $k$-NN accuracy and have higher depth RMSE, and linear accuracy also decreases slightly. Additional full-depth training therefore does not uniformly improve the grown models in this setting.

We separately shorten the Prop-1 growth horizon from 125k to 100k for MIDAS while retaining 125k total training steps and the original learning-rate schedule. This moves the final expansion from step 89,283 to step 71,425 and increases the full-depth training interval without increasing data exposure. The results are shown in \cref{tab:app_dinov3_horizon}. The cosine variant improves $k$-NN accuracy, linear accuracy, and mIoU by 0.29, 0.27, and 0.23 points, respectively, but worsens RMSE by 0.0049. Under WSD, the corresponding gains are 0.12, 0.21, and 0.53 points, with an RMSE change of only $-0.0001$. These runs use 78.0\% of fixed-depth baseline FLOPs, compared with 72.5\% for the original Prop-1 horizon.

Taken together, the DINOv3 comparisons indicate that reallocating an existing budget toward deeper stages can improve classification and segmentation, whereas extending the final stage with additional data exposure does not guarantee the same benefit. The interventions also differ in their learning-rate schedules and compute costs.

\subsubsection{Distillation from a Pretrained Teacher}
\label{app:ablation_distillation_recipes}

\begin{table}[t]
    \centering
    \caption{
        Fixed-depth DINOv3 ViT-L students distilled from the same frozen teacher. The dagger marks the recipe adopted for growing experiments. All scores are single final-layer evaluations, and three-seed results for the adopted recipe appear in \cref{tab:distillation}. \textbf{Bold} and \underline{underline} mark the best and second-best student values.}
    \label{tab:app_distill_recipes}
    \small
    \setlength{\tabcolsep}{4.2pt}
    \renewcommand{\arraystretch}{1.15}
    \begin{tabular}{lllcccc}
        \toprule
        & & & \multicolumn{2}{c}{ImageNet-1k} & ADE20K & NYU \\
        \cmidrule(lr){4-5}
        Cooldown & Peak LR & Steps
        & $k$-NN $\uparrow$ & Linear $\uparrow$
        & mIoU $\uparrow$ & RMSE $\downarrow$ \\
        \midrule
        Teacher & --- & 125k & $82.17$ & $83.49$ & $44.65$ & $0.4650$ \\
        \midrule
        none & 2e-4 & 125k & $80.78$ & $81.94$ & $43.07$ & $\mathbf{0.4618}$ \\
        none & 5e-4 & 125k & $80.46$ & $81.69$ & $41.35$ & $0.4854$ \\
        linear, 25k & 5e-4 & 125k & $81.61$ & $82.67$ & $43.10$ & $0.4751$ \\
        cosine, 25k & 5e-4 & 125k & $81.46$ & $82.55$ & $42.98$ & $0.4817$ \\
        $1-\sqrt{s}$, 25k$^{\dagger}$ & 5e-4 & 125k & $\mathbf{81.73}$ & $\mathbf{82.96}$ & $\mathbf{43.46}$ & $0.4672$ \\
        $1-\sqrt{s}$, 25k & 5e-4 & 150k & $\underline{81.69}$ & $\underline{82.89}$ & $42.76$ & $\underline{0.4638}$ \\
        $1-\sqrt{s}$, 25k & 2e-4 & 150k & $80.43$ & $81.79$ & $\underline{43.16}$ & $0.4678$ \\
        \bottomrule
    \end{tabular}
\end{table}

To select the student training recipe used in the frozen-teacher growth experiments, we compare fixed-depth DINOv3 ViT-L students distilled from the same pretrained teacher. We vary the learning rate and training duration and compare linear, cosine, and $1-\sqrt{s}$ cooldowns with training without cooldown, while holding the teacher, batch size of 2048, and stochastic-depth rate of zero fixed. Every enabled cooldown lasts 25k steps. The teacher and distillation procedure are described in \cref{app:distillation_setup}. This ablation evaluates student recipes, and the comparison of fixed-depth and grown students is reported in \cref{tab:distillation}.

The results are shown in \cref{tab:app_distill_recipes}. At $5\times10^{-4}$ and 125k steps, each cooldown improves all four metrics over training without cooldown. The $1-\sqrt{s}$ cooldown gives the strongest results among the three tested cooldown shapes: compared with no cooldown, linear accuracy increases from 81.69\% to 82.96\%, mIoU from 41.35\% to 43.46\%, and RMSE decreases from 0.4854 to 0.4672. This recipe also gives the highest $k$-NN, linear, and segmentation scores among all listed students, motivating its use for the growth experiments.

Depth estimation, however, favors a different configuration. The 125k-step student trained without cooldown at the lower peak learning rate of $2\times10^{-4}$ achieves the lowest student RMSE, 0.4618, but has weaker classification and segmentation scores. Extending the adopted $1-\sqrt{s}$ recipe to 150k steps while keeping the cooldown fixed at 25k steps improves RMSE to 0.4638, while slightly reducing both classification scores and lowering mIoU from 43.46\% to 42.76\%. The teacher remains stronger than all listed fixed-depth students on classification and segmentation.

\subsubsection{Joint Growing of MAE Encoder and Decoder}
\label{app:mae_joint_growing}

\begin{table}[t]
    \centering
    \caption{
        Encoder-only versus joint encoder--decoder growth for MAE
        ViT-L.
        Both grown models use the same Stacking Prop-1 encoder
        schedule, and joint growth additionally uses a LIDAS decoder
        schedule, as detailed in \cref{tab:app_joint_growth}.
        FLOPs include the encoder and decoder at their actual depths
        throughout training and are relative to the fixed-depth baseline.
        Results are reported as mean $\pm$ standard deviation over
        three downstream seeds, while $k$-NN is deterministic. \textbf{Bold} marks the best performance.
    }
    \label{tab:app_encdec}
    \small
    \setlength{\tabcolsep}{4.2pt}
    \renewcommand{\arraystretch}{1.15}
    \begin{tabularx}{\linewidth}{@{}Xccccc@{}}
        \toprule
        & Training & \multicolumn{2}{c}{ImageNet-1k}
        & ADE20K & NYUv2 \\
        \cmidrule(lr){3-4}
        Model & FLOPs (\%) $\downarrow$
        & $k$-NN $\uparrow$ & Linear $\uparrow$
        & mIoU $\uparrow$ & RMSE $\downarrow$ \\
        \midrule
        Baseline
        & $100.0$
        & $31.40$
        & $60.28 \pm 0.07$
        & $26.29 \pm 0.03$
        & $0.6393 \pm 0.0009$ \\
        Encoder-only growth (Stacking)
        & $79.6$
        & $33.86$
        & $\mathbf{63.64 \pm 0.08}$
        & $\mathbf{27.73 \pm 0.04}$
        & $\mathbf{0.5855 \pm 0.0001}$ \\
        Joint encoder + decoder growth (Stacking / LIDAS)
        & $73.3$
        & $\mathbf{36.30}$
        & $63.08 \pm 0.06$
        & $26.16 \pm 0.03$
        & $0.6356 \pm 0.0002$ \\
        \bottomrule
    \end{tabularx}
\end{table}

As another ablation, we examine whether growing the MAE decoder provides additional downstream benefits while saving additional compute. We compare encoder-only growth with joint encoder--decoder growth. Both configurations use MAE ViT-L, trained for 50k steps with a cosine learning-rate schedule and an effective peak learning rate of $2.4\times10^{-3}$. The encoder follows the same Stacking Prop-1 schedule in both runs. The joint variant additionally grows the decoder from two to eight layers using LIDAS, whereas the encoder-only variant retains an eight-layer decoder throughout training. Both reach the same final architecture. The growth implementation and interleaved depth schedule are described in \cref{app:growing_setup,tab:app_joint_growth}.

\cref{tab:app_encdec} shows that jointly growing the encoder and
decoder provides additional compute savings, at the cost of worse
performance on three of the four downstream metrics.
Pretraining cost decreases from 79.6\% to 73.3\% of baseline FLOPs,
a further reduction of 7.9\% relative to encoder-only growth.
Classification performance remains competitive: linear accuracy
decreases by 0.56 percentage points, from 63.64\% to 63.08\%,
while $k$-NN accuracy improves by 2.44 points, reaching 36.30\%.

The differences are larger for segmentation and depth estimation.
Joint growth reduces mIoU from 27.73 to 26.16 and increases
RMSE from 0.5855 to 0.6356, bringing both metrics close to their
fixed-depth baseline values. Thus, joint growth offers additional
compute savings while retaining strong classification results,
although encoder-only growth provides a more favorable balance
across all three downstream tasks.

\subsection{Additional Layer-wise Downstream Results}
\label{app:additional_layerwise_downstream}

\begin{figure}[t]
    \centering
    \includegraphics[width=\linewidth]{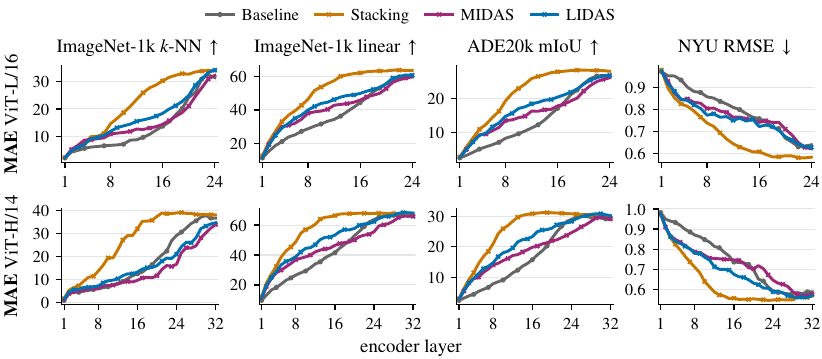}
    \caption{
        Layer-wise downstream performance across MAE growth
        strategies at 50k training steps. Each row compares the fixed-depth
        baseline with Stacking, MIDAS, and LIDAS using Prop-1.
    }
    \label{fig:app_layerwise_mae_strategies}
\end{figure}

\begin{figure}[t]
    \centering
    \includegraphics[width=\linewidth]{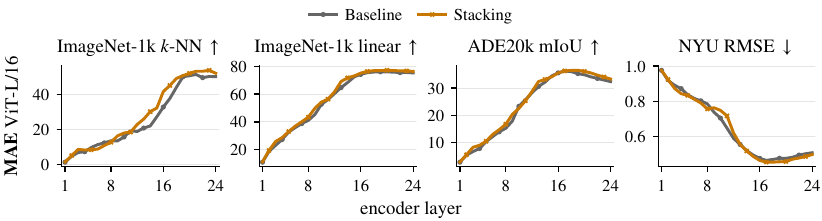}
    \caption{
        Layer-wise downstream performance of MAE ViT-L
        after 500k training steps.
        The fixed-depth baseline and Stacking model use cosine
        learning-rate schedules spanning the full training budget.
        Stacking uses Prop-1 with growth confined to the first
        50k steps and then continues training at full depth.
    }
    \label{fig:app_layerwise_mae_500k}
\end{figure}

\begin{figure}[t]
    \centering
    \includegraphics[width=\linewidth]{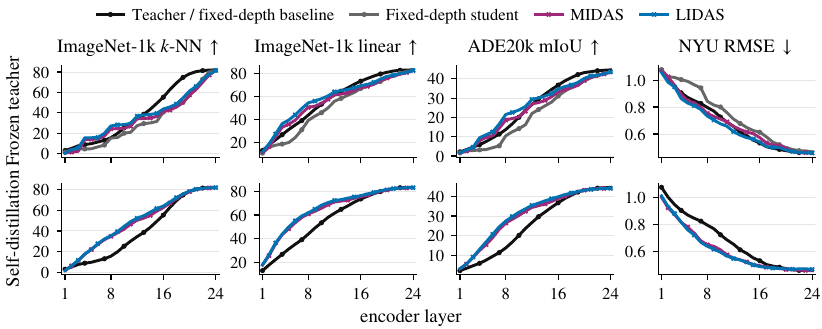}
    \caption{
        Layer-wise downstream performance of DINOv3 ViT-L
        under frozen-teacher distillation and self-distillation.
        Top: fixed-depth, MIDAS, and LIDAS students trained
        with a frozen pretrained teacher.
        Bottom: MIDAS and LIDAS trained
        under self-distillation, alongside the fixed-depth baseline.
        All runs use 125k training steps, and all grown models
        use Prop-1.
        The black curve represents the same pretrained backbone
        in both rows: the frozen teacher above and the
        fixed-depth self-distillation baseline below.
    }
    \label{fig:app_layerwise_dinov3_distill}
\end{figure}

To examine how the intermediate-layer benefits of growing depend on strategy, model scale, training duration, and teacher setting, we extend the main layer-wise analysis in \cref{subsec:exp_layerwise_downstream}. We report additional MAE strategies and model sizes, MAE after 500k training steps, and DINOv3 under frozen-teacher distillation and self-distillation. All curves follow the evaluation protocols in \cref{app:evaluation_setup}.

\textbf{MAE growing strategy and model scale.}~
\cref{fig:app_layerwise_mae_strategies} shows that the intermediate-layer improvements observed in the main MAE comparison extend to additional strategies and the larger ViT-H model, although their extent depends on the readout. Stacking gives the clearest and most sustained advantages. For ViT-L, its classification and segmentation scores approach their final-layer values around layers 16--18, while the fixed-depth baseline continues to improve substantially in later layers. For ViT-H, Stacking similarly reaches strong performance well before the final layer, with linear accuracy, segmentation, and depth estimation largely plateauing around the middle of the network.

MIDAS and LIDAS also improve intermediate linear classification and segmentation performance at both scales, with LIDAS generally giving the stronger intermediate readouts of the two. However, these advantages can narrow or reverse in later layers. For example, the ViT-L baseline catches up with MIDAS in the later layers on linear classification and segmentation. The larger model also provides an important exception in $k$-NN classification: MIDAS and LIDAS do not reproduce Stacking's broad advantage, and both fall below the baseline through much of the second half of the network. Thus, the intermediate-layer benefit is widespread but is not shared by every strategy and metric.

The ordering of strategies can also change across depth. For ViT-H, Stacking leads linear classification and segmentation across much of the middle of the network, whereas LIDAS achieves the strongest final-layer results. Several curves plateau or deteriorate slightly before the final layer, particularly for Stacking. Consequently, similar final-layer scores can conceal substantial differences in how early useful representations become accessible, and the strategy with the strongest final readout need not provide the strongest intermediate readouts.

\textbf{Extended MAE training.}~ 
At 500k steps, the broad intermediate-layer advantage of Stacking is much weaker than in the 50k-step comparison. In \cref{fig:app_layerwise_mae_500k}, the baseline and grown model follow closely aligned profiles on linear classification, segmentation, and depth estimation. The grown model does not consistently lead throughout the middle layers. For example, it has higher depth RMSE over a short interval around layers 8--11. However, the advantage does not disappear entirely. The $k$-NN curve retains a localized lead around layers 13--17, where Stacking reaches stronger performance earlier than the baseline, and the grown model retains its final-layer gains.

Both models also exhibit similar non-monotonic behavior toward the end of the network. Segmentation performance peaks around layers 16--18 and then decreases, while depth RMSE reaches its minimum around layers 17--20 before increasing toward the final layer. These patterns occur in both the baseline and Stacking. The extended-training results therefore distinguish two observations: growing can retain better final-layer performance after extended training, while its layerwise patterns become more similar to the baseline.

\textbf{DINOv3 frozen-teacher distillation and self-distillation.}~
The top row of \cref{fig:app_layerwise_dinov3_distill} shows that growing also improves many early and intermediate readouts relative to a fixed-depth student trained with the same frozen teacher. MIDAS and LIDAS generally reach stronger classification and segmentation performance earlier and reduce depth error across much of this region, although the differences narrow in later layers. LIDAS generally provides stronger early and intermediate readouts than MIDAS. The comparison with the teacher is different: the grown students can outperform it at earlier layers, but the teacher has stronger classification and segmentation readouts across much of the later network.

A distinctive feature of the frozen-teacher results is a roughly four-layer staircase, most clearly visible in $k$-NN accuracy. For the grown students, relatively small changes around layers 4--6, 8--10, and 12--14 are followed by larger gains toward the ends of the next four-layer groups. The fixed-depth student also exhibits pronounced increases near layers 8, 12, and 16. Related changes in slope appear in linear classification and segmentation, while the pattern is less clear in depth estimation and becomes less distinct in the later layers. Because this structure also appears in a student trained at fixed depth, it cannot be attributed solely to the insertion of four-layer copies during growing.

The self-distillation curves in the bottom row are smoother. Both grown models show substantial early and intermediate-layer advantages over the fixed-depth baseline, while their scores largely converge toward the final layer. These intermediate improvements remain visible even where the final-layer scores are slightly below the baseline, which further illustrates that final-layer comparisons do not fully describe the effect of growing across depth.

\subsection{Additional Computational-pattern Analyses}
\label{app:additional_computational_analyses}

We further examine how the computational changes reported in the
main text extend across growth strategies, model sizes, training
budgets, and teacher settings. We report additional attention
statistics, layer-skipping interventions, and qualitative spatial
feature maps. These comparisons also connect the computational patterns to the downstream results in \cref{app:additional_layerwise_downstream}. We use the measurements and evaluation procedures described in \cref{app:computational_analyses}.

\subsubsection{MAE: Growing Strategies, Model Scale, and Training Duration}

\begin{figure}[t]
    \centering
    \includegraphics[width=\linewidth]{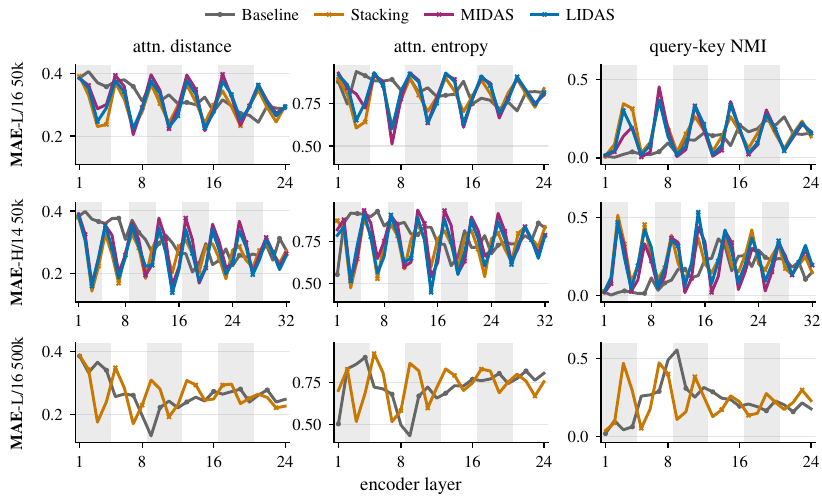}
    \caption{
        Spatial attention patterns for additional MAE configurations.
        Columns show normalized attention distance, spatial entropy,
        and query--key NMI.
        The first two rows compare the fixed-depth baseline with
        Stacking, MIDAS, and LIDAS using Prop-1 at both model sizes after 50k steps.
        The third row compares the ViT-L baseline
        and Stacking after 500k steps, with growing confined to
        the first 50k steps.
        Alternating shaded bands mark groups of four encoder layers.
    }
    \label{fig:app_attention_routing_mae_extended}
\end{figure}

\begin{figure}[t]
    \centering
    \includegraphics[width=\linewidth]{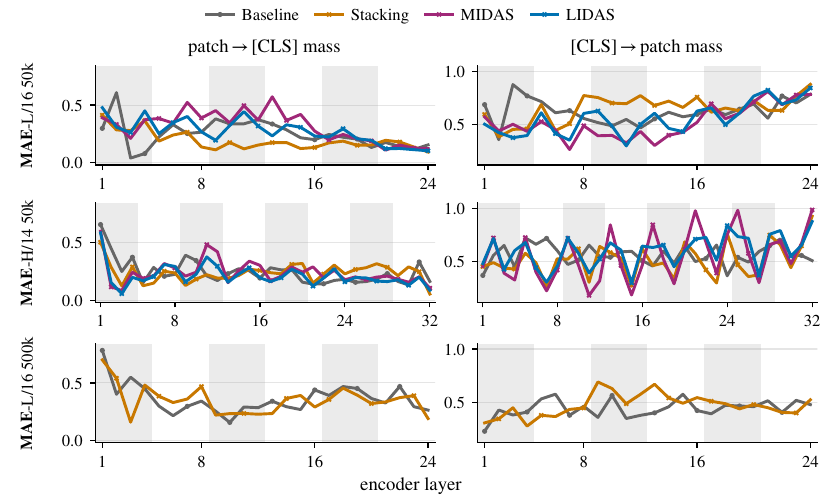}
    \caption{
        CLS--patch attention allocation for the MAE configurations
        in \cref{fig:app_attention_routing_mae_extended}.
        Columns show the average attention mass assigned by patch
        queries to CLS and the total attention mass assigned by
        the CLS query to patches.
        Rows show ViT-L at 50k steps, ViT-H at 50k steps,
        and ViT-L at 500k steps.
        Alternating shaded bands mark four-layer groups.
    }
    \label{fig:app_attention_cls_mae_extended}
\end{figure}

\textbf{Spatial attention across strategies and sizes.}~
\cref{fig:app_attention_routing_mae_extended} shows that the four-layer spatial pattern observed for Stacking in \cref{subsec:exp_computational_mae} also appears under MIDAS and LIDAS. For ViT-L at 50k steps, all three strategies repeatedly alternate between broader, higher-entropy attention and more localized, query-dependent attention. Distance and entropy generally peak near layers 5, 9, 13, 17, and 21, while NMI peaks near layers 7, 11, 15, and 19. The corresponding cycles also appear in ViT-H and continue through its additional layers. The fixed-depth baselines exhibit layer-wise variation, but lack the same regular alignment across the three statistics.

The strategies share this organization but differ in strength. For example, MIDAS has a particularly deep entropy minimum and large NMI peak near layer 7 in ViT-L. In ViT-H, LIDAS exhibits another pronounced concentration of attention near layer 15. Thus, Stacking does not consistently produce the most extreme localized or query-dependent attention. This distinction matters for interpreting the downstream results: all three strategies exhibit the spatial cycle, although their intermediate and final-layer performance differ in \cref{fig:app_layerwise_mae_strategies,tab:main_downstream}. The presence of four-layer periodicity alone therefore does not explain why one growing strategy performs better than another.

\textbf{CLS--patch attention exchange.}~
The shared spatial pattern does not imply a shared change in CLS--patch attention allocation. For ViT-L at 50k steps, \cref{fig:app_attention_cls_mae_extended} shows that Stacking reduces patch-to-CLS mass and increases CLS-to-patch mass across much of the middle of the network. MIDAS often changes these quantities in the opposite direction: patch-to-CLS mass is elevated across approximately layers 6--15, while CLS-to-patch mass is lower than the baseline through much of this region. LIDAS follows a more mixed profile. The strategies therefore differ in token interaction even when their spatial attention cycles are closely aligned.

ViT-H also shows that CLS attention need not vary smoothly. Its CLS-to-patch profiles oscillate strongly, particularly under MIDAS, with some peaks approaching one and much lower values in neighboring layers. Patch-to-CLS mass is generally less regular and does not show a consistent increase or decrease across strategies.

\textbf{Persistence after extended training.}~
At 500k steps, Stacking retains a recognizable four-layer spatial organization. Its NMI curve continues to peak near layers 3, 7, 11, 15, 19, and 23, although later peaks are generally smaller than the early ones. Distance and entropy also retain oscillations, with less regularity in the later layers. The baseline instead develops a prominent localized, low-entropy, high-NMI region around layers 8--9. At this depth, the two models can therefore occupy quite different attention regimes despite their similar downstream readouts.

This comparison connects directly to \cref{fig:app_layerwise_mae_500k}, where the broad intermediate-layer advantage of Stacking becomes smaller on linear classification, segmentation, and depth estimation. The computational distinction has not disappeared alongside that reduction in downstream differences. CLS allocation also remains different in parts of the network: Stacking assigns more CLS-to-patch attention around layers 9--14, while the profiles become closer later. These results show that distinct attention organization persists after prolonged full-depth training but with a small advantage at intermediate readouts.

\begin{figure}[t]
    \vspace{-1em}
    \centering
    \includegraphics[width=\linewidth]{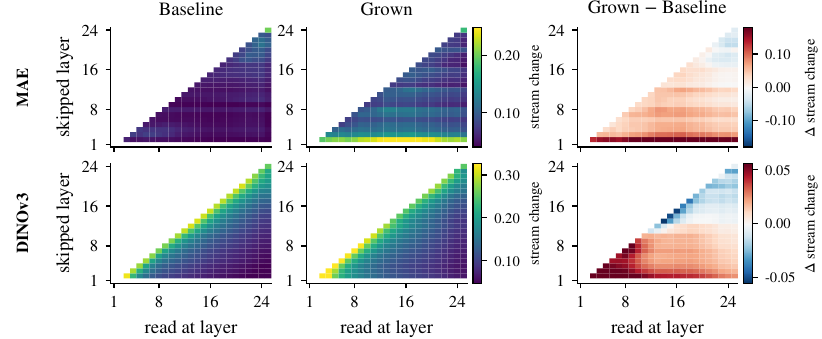}
    \caption{
        \textbf{Propagation of layer-skipping effects.}
        Cells show the relative change in the residual stream at each readout layer (horizontal axis) after skipping a block (vertical axis), compared with the unmodified model. Rows show MAE Stacking and DINOv3 LIDAS and columns show baseline, grown model, and their difference. Growing strengthens early-layer influence and weakens the effects of many late-layer interventions.
    }
    \label{fig:layer_skip}
\end{figure}

\begin{figure}[t]
    \centering
    \includegraphics[width=\linewidth]{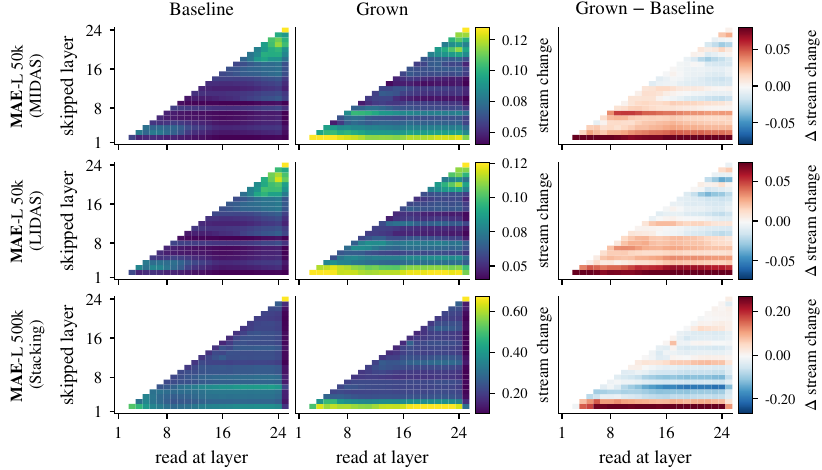}
    \caption{
        Additional layer-skipping comparisons for MAE ViT-L.
        Rows show MIDAS and LIDAS at 50k steps, followed by
        Stacking at 500k steps with a 50k growth horizon.
        All grown models use Prop-1 and cosine scheduling.
        Columns show the matched fixed-depth baseline, the grown
        model, and their difference.
        Each cell measures the relative change in the residual
        stream at the horizontal readout layer when the block
        on the vertical axis is bypassed.
        Red in the difference maps indicates larger effects
        in the grown model, and blue indicates smaller effects.
    }
    \label{fig:app_layer_skip_mae_extended}
\end{figure}

\begin{figure}[t]
    \centering
    \includegraphics[width=\linewidth]{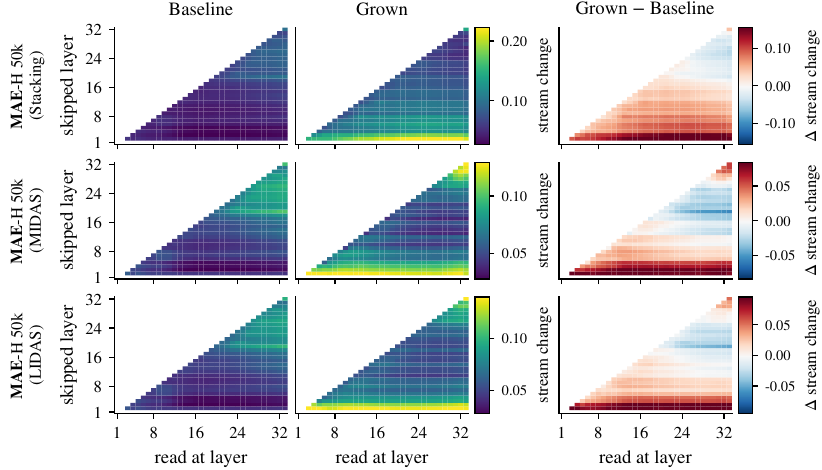}
    \caption{
        Layer-skipping comparisons for MAE ViT-H trained
        for 50k steps.
        Rows show Stacking, MIDAS, and LIDAS using Prop-1,
        each compared with the same fixed-depth baseline.
        Columns show the matched fixed-depth baseline, the grown
        model, and their difference.
        Each cell measures the relative change in the residual
        stream at the horizontal readout layer when the block
        on the vertical axis is bypassed.
        Red in the difference maps indicates larger effects
        in the grown model, and blue indicates smaller effects.
    }
    \label{fig:app_layer_skip_mae_h}
\end{figure}

\textbf{Propagation of layer-skipping effects.}~
At 50k steps, the additional strategies broadly reproduce the redistribution of intervention effects reported for ViT-L Stacking in \cref{subsec:exp_computational_mae}. In \cref{fig:app_layer_skip_mae_extended}, the earliest interventions in MIDAS and LIDAS produce larger changes across many subsequent readout layers. Several early and middle interventions also have stronger effects, whereas many later interventions have smaller effects, particularly at nearby readouts. The changes are structured across both the skipped and readout layers, rather than a uniform shift in sensitivity. The 500k-step comparison is more selective. The earliest intervention still produces a substantially larger effect in Stacking across much of the remaining network. However, several other early and middle interventions now produce smaller effects than in the baseline, while selected middle layers retain localized increases.

The ViT-H comparisons in \cref{fig:app_layer_skip_mae_h} extend this pattern to the larger-model setting. All three strategies increase the propagation of effects from the earliest layers and reduce sensitivity to many interventions in the later half of the network. For Stacking, increased effects extend broadly through interventions in approximately the first half of the model. For MIDAS and LIDAS, they are more concentrated in the early layers, with additional localized increases in the middle. This is consistent with the particularly broad intermediate-layer downstream advantage of Stacking in \cref{fig:app_layerwise_mae_strategies}, even though LIDAS achieves stronger final-layer linear classification and segmentation results. The intervention and downstream comparisons thus both show that strategy differences depend on where the network is examined.

\subsubsection{DINOv3: Growing Schedules and Teacher Settings}

\begin{figure}[t]
    \centering
    \includegraphics[width=\linewidth]{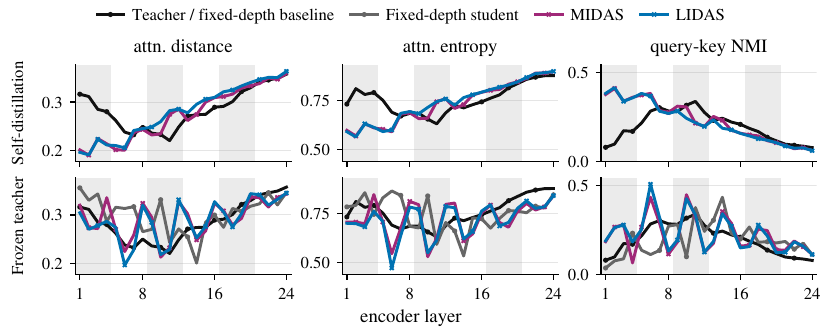}
    \caption{
        Spatial attention patterns for DINOv3 ViT-L under
        self-distillation and frozen-teacher distillation.
        Columns show normalized attention distance, spatial
        entropy, and query--key NMI.
        Top: MIDAS and LIDAS with their fixed-depth
        self-distillation baseline.
        Bottom: MIDAS, LIDAS, and fixed-depth students,
        together with their frozen pretrained teacher.
        All runs use 125k training steps, and the grown models use Prop-1.
        The black curve is the same backbone in both rows.
        Alternating shaded bands mark four-layer groups.
    }
    \label{fig:app_attention_routing_dinov3_distill}
\end{figure}

\begin{figure}[t]
    \centering
    \includegraphics[width=\linewidth]{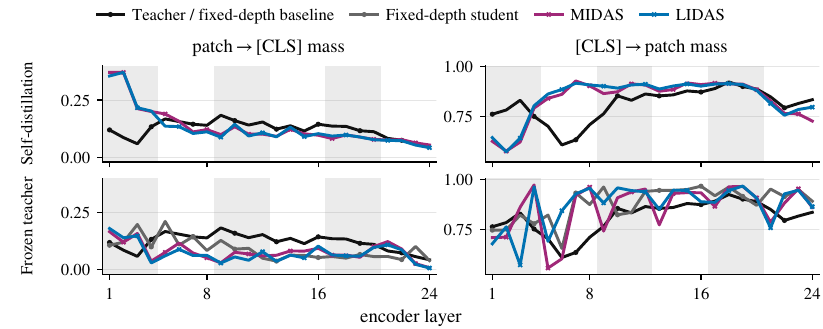}
    \caption{
        CLS--patch attention allocation for the DINOv3
        configurations in
        \cref{fig:app_attention_routing_dinov3_distill}.
        Columns show patch-to-CLS and CLS-to-patch attention mass.
        The top row shows self-distillation and the
        bottom row shows frozen-teacher distillation.
        MIDAS and LIDAS use Prop-1 in both rows.
        The fixed-depth student appears only in the bottom row, and
        the black curve represents the fixed-depth
        self-distillation baseline above and the frozen teacher
        below.
        Alternating shaded bands mark four-layer groups.
    }
    \label{fig:app_attention_cls_dinov3_distill}
\end{figure}

\textbf{Earlier attention transitions under Prop-1.}~
The self-distillation row of \cref{fig:app_attention_routing_dinov3_distill} shows that the earlier local-to-global progression reported for Prop-3 also appears under Prop-1. MIDAS and LIDAS already combine relatively short attention distance, low entropy, and high NMI in the first few layers. The baseline develops its strongest localized, query-dependent phase later, around layers 8--11. After their early localized phase, the grown models generally increase attention distance and entropy while reducing NMI, approaching the baseline's spatial statistics toward the end of the network.

The CLS-attention curves provide a corresponding early shift. Patch-to-CLS mass reaches approximately 0.35--0.37 in the first two layers of the grown models and then decreases. CLS-to-patch mass reaches its early minimum near layer 2 and recovers to approximately 0.9 around layers 6--8, ahead of the baseline's recovery. However, the final allocation differs from the main Prop-3 comparison: under Prop-1, CLS-to-patch mass falls in the last few layers and finishes below the baseline. Thus, the earlier transition is shared across growth schedules, while the late-layer attention allocation remains schedule-dependent.

This distinction complements the schedule ablations in \cref{app:ablation_growing_schedule_strategy}. The earlier attention progression therefore is generalizable across growing schedules.

\textbf{Oscillatory attention under frozen-teacher distillation.}~
For frozen-teacher distillation, similar to MAE, the grown students exhibit a more pronounced recurring spatial pattern. In the bottom row of \cref{fig:app_attention_routing_dinov3_distill}, MIDAS and LIDAS have local distance and entropy minima near layers 6, 10, 14, and 18, accompanied by NMI peaks. These are separated by broader, less query-dependent attention in neighboring layers. The contrast is strongest in the early and middle layers: for example, LIDAS reaches an NMI peak of approximately 0.5 near layer 6. The oscillations become smaller toward the end of the network.

The fixed-depth student also has substantial layer-to-layer variation, although its prominent peaks are less regularly aligned with the four-layer groups than those of the grown students. The teacher instead follows a smoother overall progression. This provides a connection to the approximately four-layer staircase in the downstream curves of \cref{fig:app_layerwise_dinov3_distill}: recurring attention regimes coexist with step-like improvements in the grown students' readouts.

CLS allocation also distinguishes the students from their teacher in \cref{fig:app_attention_cls_dinov3_distill}. Across much of the middle of the network, the grown students assign less patch-to-CLS attention than the teacher, while CLS allocates most of its attention to patches. CLS-to-patch mass varies sharply in the early layers and is frequently around 0.9 or higher later in the network. The fixed-depth student shares much of this later patch-focused CLS allocation. These differences show that grown students trained against the same frozen teacher produce different internal attention profiles.

\begin{figure}[t]
    \centering
    \includegraphics[width=\linewidth]{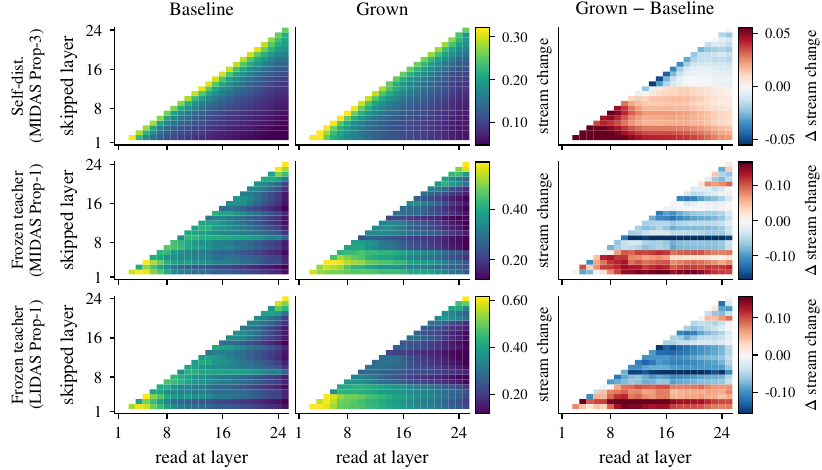}
    \caption{
        Additional layer-skipping comparisons for DINOv3 ViT-L.
        The top row compares cosine self-distillation with
        MIDAS Prop-3 against its fixed-depth baseline.
        The remaining rows compare frozen-teacher students
        grown with MIDAS or LIDAS Prop-1 against the fixed-depth
        student trained with the same teacher and recipe.
        Columns show the reference, grown model, and their
        difference in relative residual-stream change.
        The skipped block is on the vertical axis and the
        readout layer is on the horizontal axis.
        Red denotes larger effects in the grown model and blue
        denotes smaller effects.
    }
    \label{fig:app_layer_skip_dinov3_extended}
\end{figure}

\textbf{Layer-skipping under different teacher settings.}~
For self-distilled MIDAS with Prop-3, \cref{fig:app_layer_skip_dinov3_extended} shows closely matched reference and grown heatmaps, with a structured difference resembling the main LIDAS comparison. Early interventions have larger effects at subsequent middle-layer readouts, while many interventions later in the network have smaller effects, especially near the skipped block. For some interventions, the nearby response is smaller even though more distant streams change more.

The frozen-teacher comparisons exhibit more heterogeneous differences from the matched fixed-depth student. For both MIDAS and LIDAS, several early interventions produce larger effects across many later readouts. A prominent negative band near skipped layer 9 indicates a smaller effect across much of the remaining network, and many subsequent middle-layer interventions also have reduced effects.

These results extend the main observation of redistributed layer influence to both growth strategies and teacher settings. They also complement the task-dependent final results in \cref{tab:distillation}: the grown frozen-teacher students improve depth estimation over the fixed-depth student while having slightly lower linear classification accuracy. The intervention changes therefore characterize a different organization of computation.

\subsubsection{Qualitative Spatial Feature Maps}
\label{app:qualitative_feature_maps}

\begin{figure}[t]
    \centering
    \includegraphics[width=\linewidth]{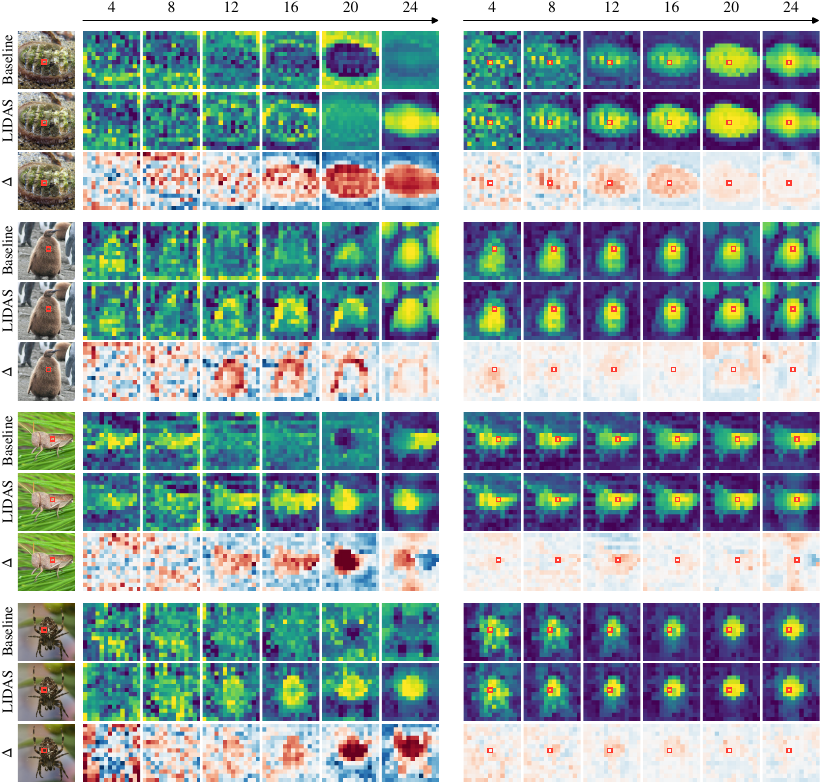}
    \caption{
        Additional spatial feature similarities across depth for DINOv3.
        Each triplet of rows shows the fixed-depth baseline, LIDAS with Prop-3, and the delta between the baseline and LIDAS for one image.
        Columns within each half show encoder layers
        4, 8, 12, 16, 20, and 24, with CLS--patch cosine similarity
        on the left and anchor-to-patch cosine similarity on the
        right. Red squares mark the anchor used for the
        patch-to-patch maps.
        LIDAS shows earlier or more persistent patterns over
        object regions, especially in the CLS--patch maps.
    }
    \label{fig:app_qualitative_dinov3}
\end{figure}

To further illustrate how growing changes spatial feature patterns across depth, \cref{fig:app_qualitative_dinov3} extends \cref{fig:qualitative_dinov3} with four additional DINOv3 examples, comparing LIDAS with Prop-3 schedule against its fixed-depth baseline. The visualization procedure is described in \cref{app:computational_analyses}. Our exploratory MAE visualizations did not reveal a consistent difference between grown and fixed-depth models, thus, we only focus on DINOv3.

In the first image, LIDAS's high patch-similarity region covers more of the object at layers 12--16, while the baseline develops a broader region around layer 20. Their late CLS--patch maps also differ, although the baseline's layer-20 map already outlines the object through lower similarity. In the second and fourth images, LIDAS develops clearer object-associated CLS--patch patterns around layers 12--16, while patch-to-patch maps are comparatively similar between models. The third image illustrates persistence: both models show object-associated CLS patterns at layers 4--8, but LIDAS maintains them more clearly through layers 12--16.

Overall, the results complement the earlier attention progression and stronger intermediate-layer readouts: LIDAS can produce earlier or more persistent spatial similarity patterns, with the clearest differences generally appearing in the CLS--patch maps.

\subsection{Robustness to Missing Visual Evidence}
\label{app:additional_results_masking}

\begin{figure}[t]
    \centering
    \includegraphics[width=\linewidth]{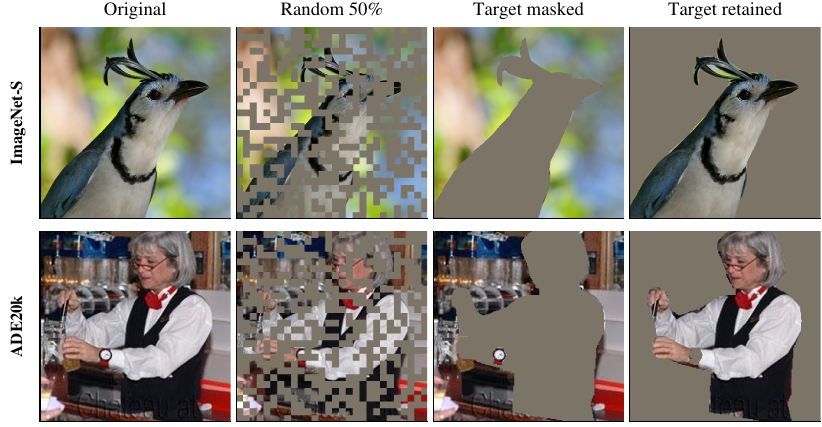}
    \caption{
        Examples of input-masking conditions. Rows show a sample from ImageNet-S and ADE20K datasets. From left to right, columns show the original image, random masking of 50\% of input patches, the target masked, and the target retained with the remaining image masked. Masked pixels are replaced with the ImageNet channel mean.
    }
    \label{fig:app_input_masking_examples}
\end{figure}

Here we examine whether the benefits of growing persist when visual evidence is removed from the input scenes. We evaluate final-layer classification and segmentation performance under random and semantic input masking for the models studied in \cref{sec:experiments}. The masking procedures and frozen-head evaluation follow \cref{app:additional_evaluations}. We report both absolute performance and retention, defined as the masked score divided by the corresponding clean score expressed as a percentage. This distinguishes performance under masking from the fraction of each model's clean performance that is preserved. An example of each masking strategy is shown in \cref{fig:app_input_masking_examples}.

\begin{figure}[t]
    \centering
    \includegraphics[width=\linewidth]{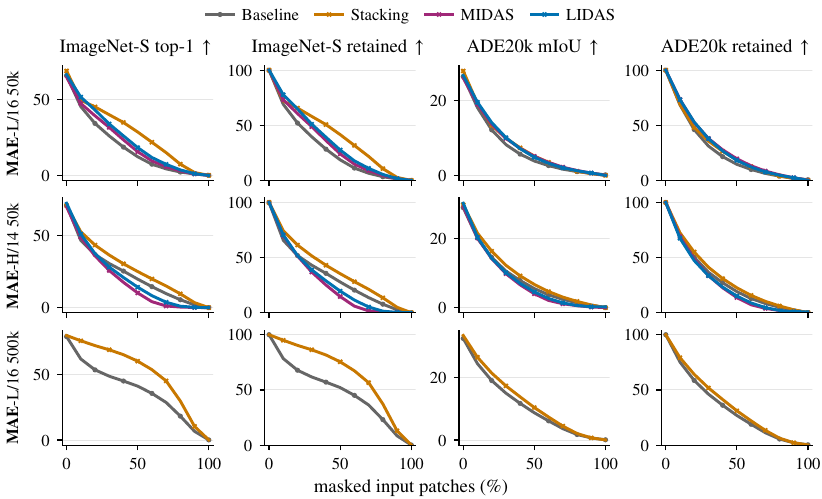}
    \caption{
        MAE robustness to random input masking.
        Rows show ViT-L at 50k steps, ViT-H at 50k steps,
        and ViT-L at 500k steps.
        All grown models use Prop-1, and the 500k-step Stacking
        run confines growing to the first 50k steps.
        Columns show ImageNet-S top-1 accuracy and its
        retention, followed by ADE20K mIoU and its retention.
        Retention is measured relative to each model's own
        clean score.
    }
    \label{fig:app_input_masking_random_mae}
\end{figure}

\begin{figure}[t]
    \centering
    \includegraphics[width=\linewidth]{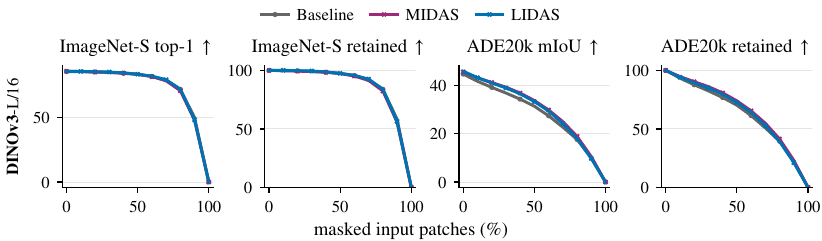}
    \caption{
        DINOv3 ViT-L robustness to random input masking.
        The fixed-depth baseline is compared with MIDAS and
        LIDAS using Prop-3.
        Columns show ImageNet-S top-1 accuracy and its
        retention, followed by ADE20K mIoU and its retention.
        Retention is measured relative to each model's own
        clean score.
    }
    \label{fig:app_input_masking_random_dinov3}
\end{figure}

\textbf{Random input masking.}~
\cref{fig:app_input_masking_random_mae} shows that the effect of growing on robustness depends on both strategy and task. For MAE ViT-L at 50k steps, Stacking provides the clearest classification advantage across moderate masking ratios, both in absolute accuracy and relative retention. MIDAS and LIDAS give smaller improvements over the baseline. For ViT-H, Stacking again preserves classification performance better, whereas MIDAS and LIDAS fall below the baseline across much of the moderate-to-high masking range. This changes the ordering observed on clean inputs: LIDAS has the highest clean classification score, but Stacking is more robust to random evidence removal.

The segmentation differences are smaller. For ViT-L at 50k steps, the grown models have broadly similar retention profiles and modest advantages over the baseline across intermediate masking ratios. For ViT-H, Stacking gives the strongest retention, while MIDAS and LIDAS generally do not improve upon the baseline after normalization by their clean scores.

The strongest classification contrast appears after 500k training steps. The clean scores of Stacking and the baseline differ by only 0.84 percentage points, yet their separation becomes much larger when patches are removed. At 50\% masking, Stacking retains approximately 75\% of its clean accuracy, compared with approximately 52\% for the baseline. Its segmentation advantage is more modest, although it also retains more performance across much of the intermediate masking range. This complements \cref{fig:app_layerwise_mae_500k}: the clean layer-wise downstream profiles become closely aligned after extended training, while a substantial difference in classification robustness remains.

DINOv3 exhibits a different pattern in \cref{fig:app_input_masking_random_dinov3}. The baseline, MIDAS, and LIDAS classification curves are closely aligned, including after normalization by clean accuracy. Segmentation shows small advantages for the grown models, but the retention curves remain similar. Classification preserves much of its performance through moderate masking ratios and drops sharply at the highest ratios, whereas segmentation deteriorates more steadily.

\begin{table}[t]
    \centering
    \caption{
        Semantic input-masking results.
        ImageNet-S reports top-1 accuracy and ADE20K reports
        whole-image mIoU.
        \emph{Target masked} replaces the selected semantic
        region with the ImageNet channel mean, and
        \emph{target retained} masks the rest of the image.
        Parentheses report retention as a percentage of the
        corresponding clean score.
        Model configurations match \cref{fig:app_input_masking_random_mae,fig:app_input_masking_random_dinov3}.
    }
    \label{tab:app_masking_semantic}
    \small
    \setlength{\tabcolsep}{4pt}
    \renewcommand{\arraystretch}{1.15}
    \begin{tabular}{@{}lcccccc@{}}
        \toprule
        & \multicolumn{3}{c}{ImageNet-S top-1 $\uparrow$}
        & \multicolumn{3}{c}{ADE20K mIoU $\uparrow$} \\
        \cmidrule(lr){2-4}
        \cmidrule(lr){5-7}
        Model
        & Clean & Target masked & Target retained
        & Clean & Target masked & Target retained \\
        \midrule
        \multicolumn{7}{l}{\textit{MAE ViT-L 50k}} \\
        \quad Baseline
        & $65.55$ & $12.26$ (19\%) & $34.20$ (52\%)
        & $26.31$ & $19.16$ (73\%) & $6.19$ (24\%) \\
        \quad Stacking
        & $\mathbf{68.63}$
        & $\mathbf{14.86}$ (22\%)
        & $\mathbf{42.58}$ (62\%)
        & $\mathbf{27.79}$
        & $\mathbf{20.76}$ (75\%)
        & $\mathbf{6.45}$ (23\%) \\
        \quad MIDAS
        & $64.60$ & $11.80$ (18\%) & $32.20$ (50\%)
        & $25.89$ & $18.92$ (73\%) & $6.02$ (23\%) \\
        \quad LIDAS
        & $66.23$ & $12.62$ (19\%) & $37.45$ (57\%)
        & $26.59$ & $19.73$ (74\%) & $6.31$ (24\%) \\
        \midrule
        \multicolumn{7}{l}{\textit{MAE ViT-H 50k}} \\
        \quad Baseline
        & $71.14$ & $13.63$ (19\%) & $42.12$ (59\%)
        & $28.90$ & $21.82$ (76\%) & $6.55$ (23\%) \\
        \quad Stacking
        & $70.83$
        & $\mathbf{15.83}$ (22\%)
        & $\mathbf{48.12}$ (68\%)
        & $29.73$
        & $\mathbf{22.96}$ (77\%)
        & $\mathbf{6.74}$ (23\%) \\
        \quad MIDAS
        & $69.98$ & $14.17$ (20\%) & $32.77$ (47\%)
        & $29.21$ & $21.58$ (74\%) & $6.31$ (22\%) \\
        \quad LIDAS
        & $\mathbf{72.49}$
        & $15.76$ (22\%)
        & $45.02$ (62\%)
        & $\mathbf{30.19}$
        & $22.76$ (75\%)
        & $\mathbf{6.74}$ (22\%) \\
        \midrule
        \multicolumn{7}{l}{\textit{MAE ViT-L 500k}} \\
        \quad Baseline
        & $78.82$ & $15.40$ (20\%) & $57.80$ (73\%)
        & $32.37$ & $24.80$ (77\%) & $6.49$ (20\%) \\
        \quad Stacking
        & $\mathbf{79.66}$
        & $\mathbf{17.87}$ (22\%)
        & $\mathbf{63.70}$ (80\%)
        & $\mathbf{33.32}$
        & $\mathbf{25.78}$ (77\%)
        & $\mathbf{6.60}$ (20\%) \\
        \midrule
        \multicolumn{7}{l}{\textit{DINOv3 ViT-L}} \\
        \quad Baseline
        & $85.30$ & $33.20$ (39\%) & $79.39$ (93\%)
        & $44.65$ & $34.43$ (77\%) & $11.82$ (26\%) \\
        \quad MIDAS
        & $\mathbf{85.54}$
        & $36.04$ (42\%)
        & $80.76$ (94\%)
        & $45.57$
        & $38.27$ (84\%)
        & $\mathbf{13.52}$ (30\%) \\
        \quad LIDAS
        & $85.38$
        & $\mathbf{37.12}$ (43\%)
        & $\mathbf{80.93}$ (95\%)
        & $\mathbf{45.83}$
        & $\mathbf{38.72}$ (84\%)
        & $13.30$ (29\%) \\
        \bottomrule
    \end{tabular}
\end{table}

\textbf{Semantic input masking.}~
\cref{fig:app_input_masking_examples} illustrates the distinction between removing scattered patches and removing a selected semantic region. For ImageNet-S, the target is the annotated foreground. For ADE20K, the target is the largest eligible class region in the image. The semantic interventions therefore preserve different parts of the scene, whose usefulness also depends on the downstream task.

On ImageNet-S, all models perform substantially better when the target is retained than when it is masked. Stacking improves both conditions across the three MAE comparisons in \cref{tab:app_masking_semantic}. Its largest classification gains occur when the target is retained: for ViT-L at 50k steps, accuracy increases from 34.20\% to 42.58\%, with retention rising from 52\% to 62\%. For ViT-H, Stacking reaches 48.12\% against 42.12\% for the baseline, despite having slightly lower clean accuracy. The advantage persists at 500k steps, where Stacking reaches 63.70\% against 57.80\%, retaining 80\% rather than 73\% of clean accuracy. These results extend its classification robustness advantage beyond random masking.

The other MAE strategies give mixed results. LIDAS improves target-retained classification at both sizes, while MIDAS performs worse than the corresponding baseline. The ViT-H MIDAS decrease is particularly pronounced: target-retained accuracy falls from 42.12\% to 32.77\%, a much larger difference than its 1.16-point clean-accuracy deficit. For ADE20K, the MAE retention differences are generally smaller. Target masking preserves approximately 73--77\% of clean whole-image mIoU, whereas retaining only the target preserves approximately 20--24\%.

DINOv3 shows a clearer benefit from growing under semantic masking than under random masking. With the ImageNet-S target masked, MIDAS and LIDAS increase accuracy from 33.20\% to 36.04\% and 37.12\%, respectively. Both also improve target-retained accuracy. The largest segmentation gains occur with the ADE20K target-masked: mIoU increases from 34.43 to 38.27 for MIDAS and 38.72 for LIDAS, gains of 3.84 and 4.29 points. These exceed their clean-score advantages of 0.92 and 1.18 points, with retention increasing from 77\% to 84\% for both strategies. Both also improve the target-retained condition, reaching 13.52 and 13.30 mIoU against 11.82 for the baseline.

Overall, these results support improved robustness to missing visual evidence as an additional benefit of growing. The clearest gains appear in MAE Stacking's classification results across both masking types and in grown DINOv3 under semantic masking. Although the benefits vary across strategies, tasks, and interventions, the improvements in relative retention show that they are not explained solely by higher clean scores. Growing can therefore improve both downstream performance on complete inputs and the ability to maintain that performance when visual evidence is removed.

\subsection{MAE Adaptation through Fine-tuning and Larger Heads}
\label{app:mae_adaptation}

To examine whether the benefits of growing depend on a linear readout or persist under task-specific adaptation, we compare linear and nonlinear probes for ImageNet classification task on frozen MAE representations and additionally fine-tune the full MAE encoder following the original recipe. The probe and fine-tuning procedures are described in \cref{app:additional_evaluations}, with the classification protocols in \cref{app:eval_classification}. The probe analyses assess how effectively different heads can use the pretrained representations, whereas fine-tuning evaluates the performance obtained after adapting the representations themselves.

\begin{table}[t]
    \centering
    \caption{
        MAE adaptation on ImageNet-1k, measured by top-1
        accuracy (\%).
        $k$-NN uses frozen final-layer CLS features,
        linear and two- or three-layer MLP probes use frozen
        final-layer CLS and mean-pooled patch features, and
        fine-tuning (denoted by FT) updates the full encoder. The results are shown for a single seed.
        All grown models use Prop-1, with growth horizon
        of 50k steps for the 500k-step runs.
        \textbf{Bold} marks the highest accuracy per column
        within each block.
    }
    \label{tab:app_mae_heads}
    \small
    \setlength{\tabcolsep}{5pt}
    \renewcommand{\arraystretch}{1.15}
    \begin{tabular}{@{}lccccc@{}}
        \toprule
        Model
        & $k$-NN & Linear & MLP-2 & MLP-3 & FT \\
        \midrule
        \multicolumn{6}{l}{
            \textit{MAE ViT-L, 50k steps}
        } \\
        \quad Baseline
        & $31.40$ & $60.28$ & $61.71$ & $62.01$ & $82.24$ \\
        \quad Stacking
        & $\mathbf{33.86}$
        & $\mathbf{63.70}$
        & $\mathbf{64.58}$
        & $\mathbf{64.85}$
        & $\mathbf{83.77}$ \\
        \midrule
        \multicolumn{6}{l}{
            \textit{MAE ViT-L, 500k steps}
        } \\
        \quad Baseline
        & $49.96$ & $75.18$ & $75.41$ & $74.81$
        & $\mathbf{85.71}$ \\
        \quad Stacking
        & $\mathbf{51.86}$
        & $\mathbf{76.29}$
        & $\mathbf{76.38}$
        & $\mathbf{75.71}$
        & $85.62$ \\
        \midrule
        \multicolumn{6}{l}{
            \textit{MAE ViT-H, 50k steps}
        } \\
        \quad Baseline
        & $36.59$ & $66.39$ & $66.88$ & $66.31$ & $84.64$ \\
        \quad Stacking
        & $\mathbf{37.99}$ & $66.16$ & $64.92$ & $66.39$
        & $\mathbf{85.14}$ \\
        \quad MIDAS
        & $33.83$ & $65.86$ & $66.18$ & $66.22$ & $84.39$ \\
        \quad LIDAS
        & $34.60$
        & $\mathbf{68.10}$
        & $\mathbf{68.00}$
        & $\mathbf{68.01}$
        & $84.91$ \\
        \midrule
        \multicolumn{6}{l}{
            \textit{MAE ViT-H, 500k steps}
        } \\
        \quad Baseline
        & $31.06$ & $75.12$ & $74.38$ & $73.82$ & $86.79$ \\
        \quad Stacking
        & $31.90$ & $72.40$ & $71.66$ & $71.31$ & $86.81$ \\
        \quad MIDAS
        & $40.28$ & $76.53$ & $76.05$ & $75.35$ & $86.80$ \\
        \quad LIDAS
        & $\mathbf{44.59}$
        & $\mathbf{78.03}$
        & $\mathbf{77.70}$
        & $\mathbf{76.76}$
        & $\mathbf{86.84}$ \\
        \bottomrule
    \end{tabular}
\end{table}

\textbf{Nonlinear probes on frozen representations.}~
\cref{tab:app_mae_heads} shows that the advantages observed with linear probing generally persist with nonlinear MLP heads with more capacity. For MAE ViT-L at 50k steps, Stacking improves linear accuracy by 3.42 percentage points over the baseline. Both models benefit from the MLP probes, but Stacking retains gains of 2.87 points with MLP-2 and 2.84 points with MLP-3. At 500k steps, its advantages are smaller but remain present across all frozen readouts, including gains of 0.97 and 0.90 points with the two MLP heads. The benefit of growing therefore persists when the linear classifier is replaced with a nonlinear readout.

For MAE ViT-H, LIDAS achieves the strongest linear and MLP results at both training budgets. At 500k steps, its improvements over the baseline are 2.91 points with the linear head, 3.32 points with MLP-2, and 2.94 points with MLP-3. Conversely, the weaker frozen-readout performance of Stacking at this budget is not recovered by increasing head depth: its deficit remains approximately 2.5--2.7 points across the three trained probes. Thus, nonlinear heads largely preserve the strategy differences observed with linear probing.

Increasing head capacity does not, however, monotonically improve accuracy under the tested recipes. At 500k steps, MLP-3 falls below the linear probe for every listed model, and both MLP heads fall below the linear probe for all MAE ViT-H configurations.

\textbf{End-to-end fine-tuning.}~
After 50k pretraining steps, growing can also provide a stronger initialization for full fine-tuning. MAE ViT-L Stacking improves from the baseline's 82.24\% to 83.77\%, a gain of approximately 1.5 percentage points. For MAE ViT-H, Stacking reaches 85.14\% against 84.64\% for the baseline, while LIDAS reaches 84.91\% and MIDAS remains slightly below the baseline at 84.39\%. The strategy ranking differs from that of the trained probes with frozen backbone: LIDAS leads those readouts, whereas Stacking obtains the highest fine-tuned score.

After 500k pretraining steps, the fine-tuned results become closely matched despite the remaining differences between frozen readouts. For MAE ViT-L, Stacking retains an advantage with every frozen evaluation but finishes slightly below the baseline after fine-tuning, at 85.62\% versus 85.71\%. The contrast is particularly clear for MAE ViT-H: linear-probe accuracy ranges from 72.40\% to 78.03\% across the listed models, while all fine-tuned scores lie between 86.79\% and 86.84\%. Substantial differences in the usefulness of frozen features can therefore coexist with nearly identical performance after full adaptation.

These results show that the advantages of growing extend to higher-capacity downstream heads and can transfer to end-to-end fine-tuning, with the clearest fine-tuning gains after shorter pretraining. With longer pretraining, benefits remain visible in frozen-feature evaluations, while grown models achieve fine-tuned performance comparable to their fixed-depth counterparts.

\subsection{Representation and Weight Similarity}
\label{app:representation_weight_similarity}

To examine how growing shapes representations and parameters across depth, and whether its effects remain after longer pretraining, we compare within-model cosine similarity between layer representations and between consecutive four-layer groups of weights. These analyses complement the earlier downstream and attention results. Full implementation details are provided in \cref{app:additional_evaluations}.

\begin{figure}[t]
    \centering
    \includegraphics[width=\linewidth]{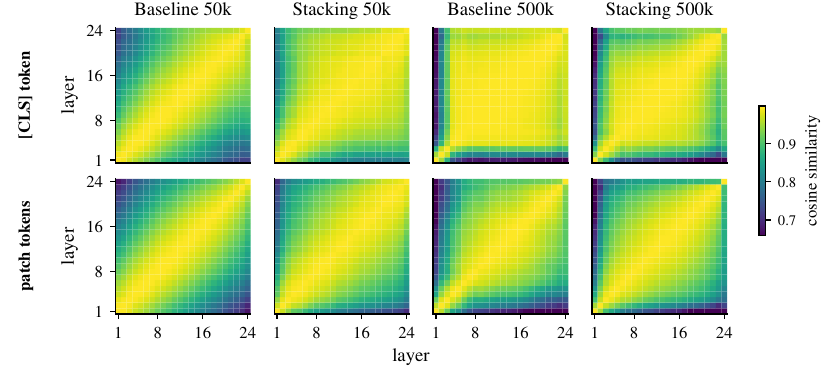}
    \caption{
        Layer-wise representation similarity between MAE ViT-L features for the fixed-depth baseline and Stacking at 50k and 500k training steps. Rows show similarity for CLS-only and patch tokens and columns show the model and training step. At 50k, Stacking has broader intermediate-layer CLS alignment. At 500k, both models exhibit strong intermediate alignment, while differences on the final CLS representations remain visible.
    }
    \label{fig:app_representation_similarity_mae}
\end{figure}

\begin{figure}[t]
    \centering
    \includegraphics[width=\linewidth]{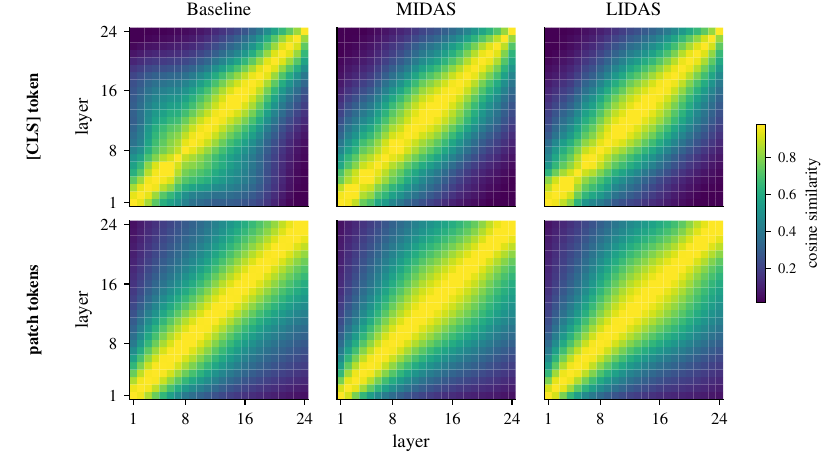}
    \caption{
        Layer-wise representation similarity in DINOv3 for the fixed-depth baseline, MIDAS, and LIDAS. Rows show similarity CLS-only and patch tokens. The three models retain a similar overall progression across depth, with localized differences in alignment. Patch-token similarity extends over a broader depth range than CLS similarity.
    }
    \label{fig:app_representation_similarity_dinov3}
\end{figure}

\textbf{Representation similarity.}~
For MAE at 50k steps, Stacking exhibits stronger alignment across intermediate CLS representations than the fixed-depth baseline (\cref{fig:app_representation_similarity_mae}). At 500k, both models show broad intermediate-layer alignment, although Stacking's final CLS representations remain more distinct from its middle layers. Differences between patch-token similarity patterns are smaller. For DINOv3, the baseline, MIDAS, and LIDAS retain broadly similar patterns, with alignment generally decreasing as layers become farther apart (\cref{fig:app_representation_similarity_dinov3}). Growing introduces localized changes without substantially altering this overall structure.

\begin{figure}[t]
    \centering
    \includegraphics[width=\linewidth]{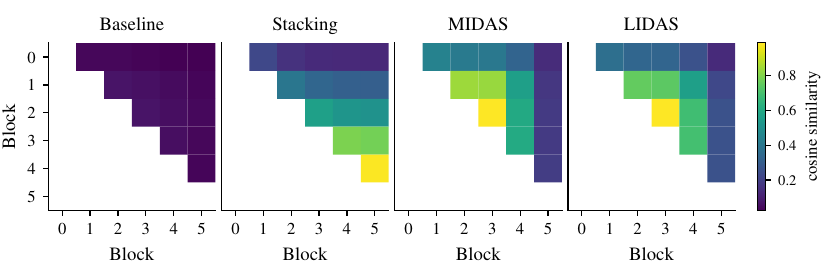}
    \caption{
        Weight similarity between consecutive
        four-layer groups in MAE ViT-L for the fixed-depth baseline, Stacking, MIDAS, and LIDAS. The axes labeled \emph{Block} index groups 0--5, corresponding to encoder layers 1--4 through 21--24. The strongest alignment occurs between the final groups for Stacking and between the central groups for MIDAS and LIDAS, matching the locations of their final source--copy pairs.
    }
    \label{fig:app_weight_similarity_strategies_mae}
\end{figure}

\begin{figure}[t]
    \centering
    \includegraphics[width=\linewidth]{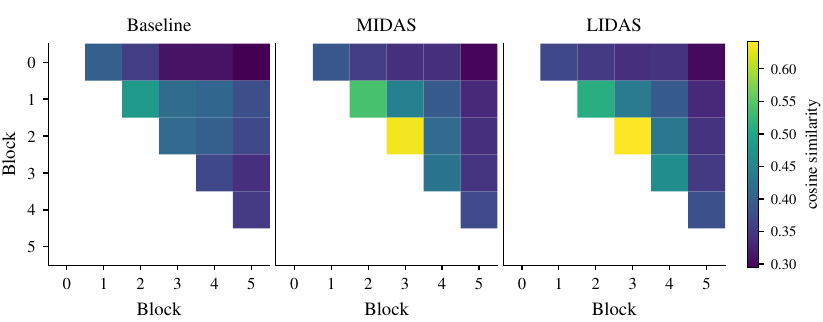}
    \caption{
        Weight similarity between consecutive four-layer groups in DINOv3 for the fixed-depth baseline, MIDAS, and LIDAS. Group indices follow \cref{fig:app_weight_similarity_strategies_mae}. Both grown models exhibit their strongest alignments in the middle blocks.
    }
    \label{fig:app_weight_similarity_strategies_dinov3}
\end{figure}

\textbf{Weight similarity across strategies.}~
The weight similarities show a clearer signature of the copying strategy (\cref{fig:app_weight_similarity_strategies_mae,fig:app_weight_similarity_strategies_dinov3}). Indices 0--5 denote consecutive groups of four layers. In MAE, Stacking has its strongest alignment between the final groups 4 and 5, whereas MIDAS and LIDAS peak between the central groups 2 and 3. DINOv3 exhibits the same central peak for MIDAS and LIDAS. These locations match the source--copy pairs at the final growth event, showing that the parameters retain a visible signature of where copying occurred.

\begin{figure}[t]
    \centering
    \includegraphics[width=\linewidth]{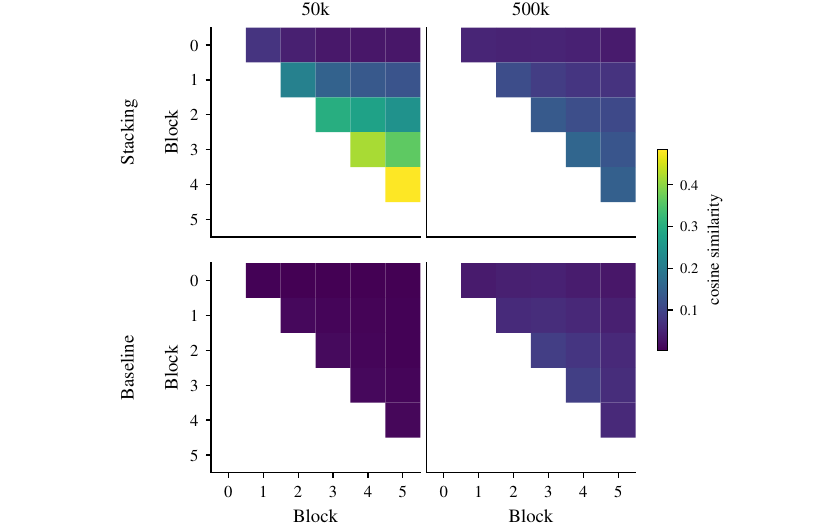}
    \caption{
        MAE weight similarity at 50k and 500k training steps.
        The top row shows Stacking and the bottom row the
        fixed-depth baseline, and the left and right columns show
        50k and 500k steps, respectively.
        Stacking's strong alignment among later groups is
        substantially weaker at 500k, although some structure
        remains relative to the baseline.
    }
    \label{fig:app_weight_similarity_mae}
\end{figure}

\textbf{Extended MAE training.}~
In the extended MAE configuration, Stacking's strong alignment among later groups is substantially weaker than at 50k, while the baseline shows a modest increase in similarity (\cref{fig:app_weight_similarity_mae}). Some differences remain, but the contrast between the models is less pronounced and the similarity patterns of the grown model become very similar to the baseline.

These comparisons show that growing leaves a clear signature of the copying strategy in the weights, while changes in representation similarity are more dependent on the model and token type. In MAE, the strongest weight alignments weaken with longer pretraining, while recurring attention patterns remain visible. This suggests that growing can establish lasting patterns of computation (\cref{app:additional_computational_analyses}) even as copied parameters become less closely aligned.

\end{document}